\documentclass[10pt,twocolumn,letterpaper]{article}
\pdfoutput=1
\usepackage{cvpr}
\usepackage{times}
\usepackage{epsfig}
\usepackage{graphicx}
\usepackage{amsmath}
\usepackage{amssymb}

\usepackage{fancyhdr}
\usepackage{bm}
\usepackage{amsbsy}
\usepackage{mathtools}
\usepackage[export]{adjustbox}
\usepackage{framed}
\usepackage{mdframed}

\DeclareMathOperator*{\argmax}{arg\,max}
\usepackage{multirow}
\usepackage{caption} %this one prevents altering figure captions
\usepackage[breaklinks=true,bookmarks=false]{hyperref}

\cvprfinalcopy % *** Uncomment this line for the final submission

\def\cvprPaperID{****} % *** Enter the CVPR Paper ID here

{%
\setlength{\fboxsep}{1ex}%
\setlength{\fboxrule}{1pt}%
}%

\begin{document}

\fancypagestyle{plain}{
\fancyhead[C]{Pattern Recognition, 2016 - To Appear}
\fancyhead[R]{} 
\fancyhead[L]{}}

%%%%%%%%% TITLE
\title{Multi-scale Volumes for Deep Object Detection and Localization}

\author{Eshed Ohn-Bar and Mohan M. Trivedi\\
University of California, San Diego\\
La Jolla, CA 92093-0434\\
{\tt\small \{eohnbar,mtrivedi\}@ucsd.edu}
}

\maketitle
%\thispagestyle{empty}

%%%%%%%%% ABSTRACT
\begin{abstract}
   This study aims to analyze the benefits of improved multi-scale reasoning for object detection and localization with deep convolutional neural networks. To that end, an efficient and general object detection framework which operates on scale volumes of a deep feature pyramid is proposed. In contrast to the proposed approach, most current state-of-the-art object detectors operate on a single-scale in training, while testing involves independent evaluation across scales. One benefit of the proposed approach is in better capturing of multi-scale contextual information, resulting in significant gains in both detection performance and localization quality of objects on the PASCAL VOC dataset and a multi-view highway vehicles dataset. The joint detection and localization scale-specific models are shown to especially benefit detection of challenging object categories which exhibit large scale variation as well as detection of small objects. 
\end{abstract}

\section{Introduction}
\begin{figure}[!t]
\centering
\begin{tabular}{c}
\includegraphics[width=3in]{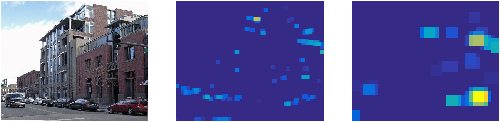}
\\
\includegraphics[width=3in]{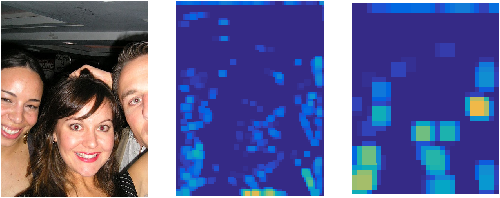}
\\
\includegraphics[width=3in]{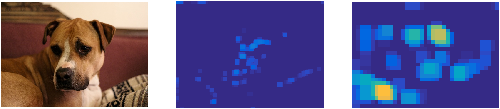}
 \end{tabular}
 \caption{Convolutional feature responses at different image scales of two octaves apart. Different feature channels are visualized for each image. The responses are scale-selective, capturing different levels of contextual information. This phenomenon is studied and modeled in this work using scale volumes in order to obtain better object detection and localization performance.}
 \label{fig:pyrvis}
\end{figure}

Visual recognition with computer vision has been rapidly improving due to the modern deep Convolutional Neural Network (CNN). The current success is fueled by large datasets, with pre-training of the network for a supervised object classification task on a large dataset \cite{Krizhevsky}, and consequent adaptation for new tasks such as object detection \cite{overfeat,rcnn} or scene analysis \cite{Shi2016448,Zuo:2015:EBD:2796563.2796624}. The success of CNNs is attributed to the rich representation power of the deep network. Therefore, much of the current research is concentrated on better understanding properties captured by CNN representations. When transferring the network from a classification task to a detection and localization task, performance is greatly influenced by the ability to capture contextual and multi-scale information \cite{holdingback}. The main aim of this study is in the evaluation and improvement of this ability for CNNs using better multi-scale feature reasoning. 

\begin{figure*}[!t]
\centering
\begin{tabular}{cc}
\begin{tabular}{c}
 \raisebox{-30ex}{\includegraphics[trim = 10mm 120mm 219mm 3mm, clip,width=0.5in]{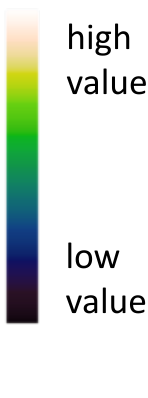}}
 \end{tabular}
\begin{tabular}{c}
   \includegraphics[trim = 10mm 25mm 2mm 0mm, clip,width=5in]{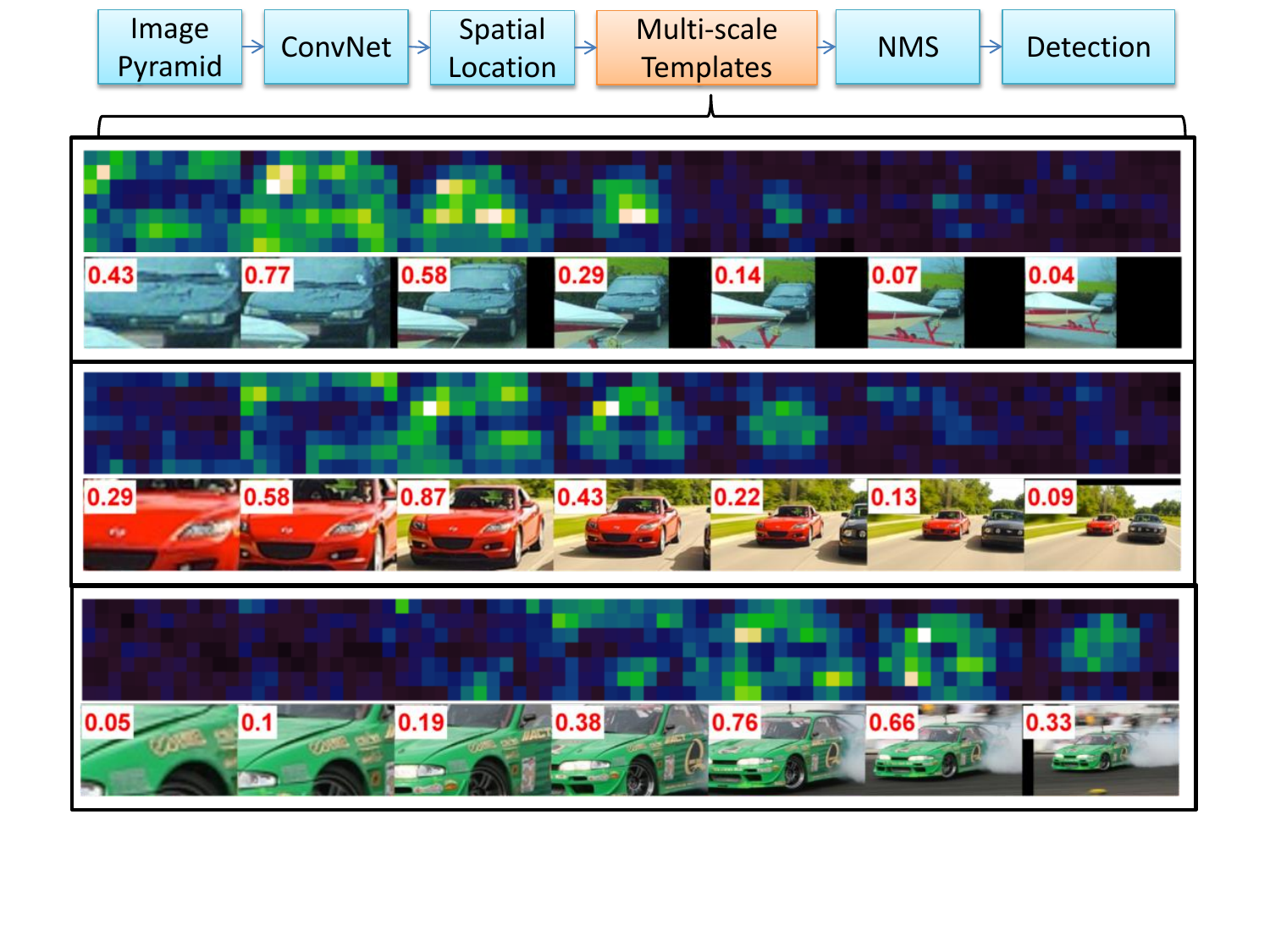}
 \end{tabular}
  \end{tabular}
 \caption{Pipeline of the proposed multi-scale structure (\textbf{MSS}) approach for studying the role of contextual and multi-scale cues in object detection and localization. Examples of some of the learned MSS models for `car' over CNN features are shown, with brighter colors implying greater discriminative value. In {\color{red}red} text is the overlap of the annotated ground truth object with a fixed model size. Note how each MSS template selects discriminative information across multiple scales, such as road and part information. 
  }
 \label{fig:motiv1}
\end{figure*}

The biological vision system can recognize and locate objects under wide variability due in part to contextual reasoning. This is of particular importance when different image and object scales are considered. Hence, the tasks of capturing contextual cues and modeling multi-scale information are interleaved. Take for instance a car detection task as depicted in Fig. \ref{fig:pyrvis}. Contextual reasoning appears at different image scales and spatial locations, from fine-grained part information (e.g. bumper, license plate, or tail lights occurring at certain configurations w.r.t. object orientation) and up to contextual scene cues such as road cues or relationship to other objects. Fig. \ref{fig:pyrvis} depicts convolutional feature responses computed at twice and half the original image size for a selected feature channel. As can be seen, the responses differ both in magnitude and location depending on the image scale. Responses at different scales contain relevant contextual information for detection and localization. It has been known that CNNs can capture increasingly semantic representations at each layer \cite{netvis}, yet detection performance varies greatly w.r.t. appearance variations (scale, orientation, occlusion, and truncation) \cite{holdingback}. Therefore, contextual multi-scale information can help resolve such challenging cases. This work aims to analyze the benefit of training models that pool features over multiple image scales, both at adjacent and remote scales, on object detection (Fig. \ref{fig:motiv1}). Furthermore, the inference label space is adjusted to better leverage contextual multi-scale information in the localization of objects.  %This paper is focused on studying the role of multi-scale cues and reasoning on object detection and localization with CNNs. 
%inline with other related studies leveraging multi-scale feature responses \cite{dollar2015fast,mscnn,yannmssseg,hed}. 
%This phenomenon is studied and modeled in this work, 
%The approach allows for a straight-forward integration of context and analysis of the impact of context in CNN-based object detection.

	\subsection{Contributions}
	The main contributions presented in this work are as follows:

\begin{enumerate}

\item Multi-scale framework: we propose a framework for understanding CNN responses at multiple image scales. By training models that learn to pool features across multiple scales and appropriately designing the inference label space, the proposed framework is used to perform novel analysis useful in obtaining insight into the role of multi-scale and contextual information. In particular, the impact of dataset size and properties, impact of different scales and object properties, types of detection and localization errors, and model visualization are addressed. The framework generalizes current state-of-the-art object detectors which perform single-scale training and independent model testing across scales.

\item Better detection and localization: Replacing the commonly used local region classification pipeline for detection with a proposed set of joint detection and localization, scale-specific, context-aware, multi-scale volume models is shown to improve detection and localization quality. The contextual information is shown to be particularly useful in resolving challenging objects, such as objects at small scale. Experimental results demonstrate generalization of the proposed, \textbf{multi-scale structure (MSS)}, approach across feature types (CNN or hand-designed features) and datasets. The approach is light-weight in memory and computation, and is therefore useful for a variety of application domains requiring a balance between robust object detection and computational cost. 

\end{enumerate}

\begin{figure*}[!t]
\centering
\resizebox{17.5cm}{!} {
\begin{tabular}{ccc}
  \includegraphics[trim = 0mm 103mm 99mm 0mm, clip,width=2.2in]{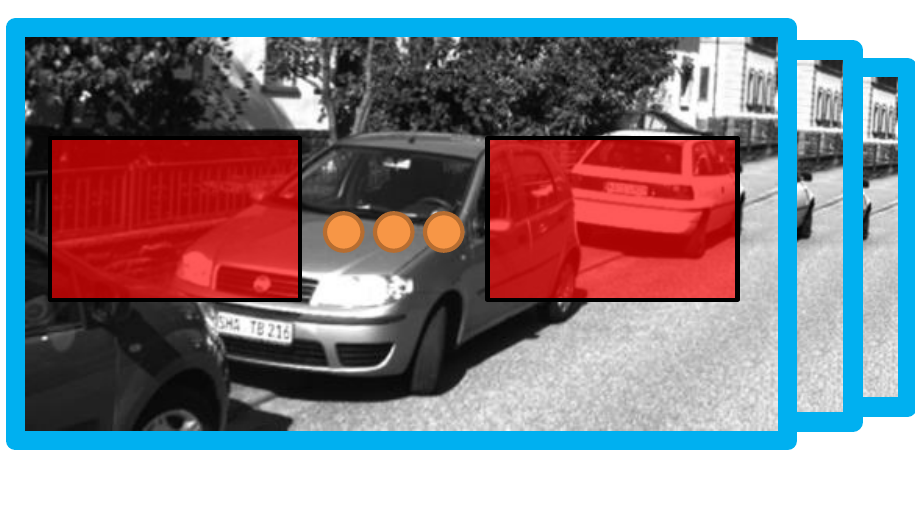}
  &
 \raisebox{+3ex}{\includegraphics[trim = 0mm 20mm 95mm 100mm, clip,width=2.2in]{traditionalvis.pdf}}
  &
   \includegraphics[trim = 0mm 103mm 101mm 0mm, clip,width=2.2in]{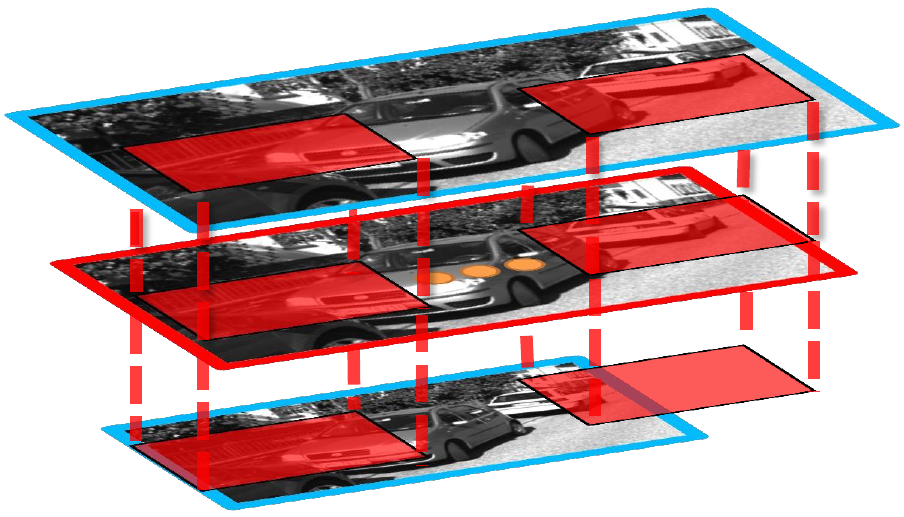}
 \\
\multicolumn{2}{c}{(a) Traditional, single-scale approaches (image pyramid and template pyramid)} & (b) Our proposed approach
\\
  \end{tabular}
  }
   \caption{Traditional approaches are limited in ability to capture contextual cues due to a single-scale training and testing of a single-scale local region. The proposed Multi-Scale Structure (MSS) approach extends the local regions across scales of an image pyramid to operate on scale volumes. The inference label space is modified as well to predict a localization label. The access to scale volumes across all scales of the image pyramid in training and testing time allows visualizing contextual cues and analyzing their role in detection and localization.
   }
 \label{fig:main}
\end{figure*}

\section{Related Research Studies}

This study aims to better understand the benefits of improved multi-scale reasoning for object detection and localization. To that end, deep features are extracted at multiple image scales, and models that can perform inference over scale volumes and leverage contextual cues across different scales are trained. The analysis provided by such models is complementary to existing related research studies discussing schemes for object detection and localization with multi-scale, contextual, and deep architectures, as will be discussed below.

\textbf{Multi-scale detection}: Traditional multi-scale object detection schemes employ a sliding window, which is a local, fixed-sized region in the image. The local region is scored in a classification task for an object presence, a process done exhaustively over different image locations and at different scales. In training, all training samples are re-sized to a fixed template size, thereby removing any scale-specific information and resulting in a single-scale model. In test time, local regions are classified independently across locations and scales. This limits the model's ability to well-localize an object and capture contextual cues. For instance, the example images in Fig. \ref{fig:motiv1} would be scored independently, despite the highly structured information across scales. Finally, resolving multiple detections is handled with a heuristic Non-Maxima Suppression (NMS) module, which has no access to the image evidence. Several works have challenged this widely used pipeline. This includes the works of \cite{parkmultires,contextboost,ohnbarTITS15,peds100rodrigo,rakesh}, which consider training multiple-resolution models. Such techniques were proposed for better handling appearance variation due to scale. The multi-resolution framework of \cite{resolutionSpeed} involves rejecting windows at low resolutions before the rest of the image pyramid is processed, thereby achieving speed gains. As the models trained in the aforementioned studies are still single-scale models, testing involves scoring each image location and scale. The contrast between the aforementioned studies and this work is that we incorporate a scale localization label into the label space of the detector, and consequently train models that operate on all scales of the image pyramid (see Fig. \ref{fig:main} for a high-level contrast). The approach explicitly accounts for variation in appearance due to scale and incorporates contextual cues for better localization. We note that the impact on detection and localization quality due to employing features at all scales, both remote and adjacent, has been rarely studied in related studies. As will be shown in Section \ref{sec:MSS}, the studied framework generalizes the studies of \cite{parkmultires,contextboost,ohnbarTITS15,peds100rodrigo} which do not modify the multi-scale sliding window pipeline, and therefore provides complementary analysis.

\textbf{CNN-based object detection}: CNNs are a long-studied class of models \cite{fukushima1980neocognitron,rumelhart1985learning,rumelhart1988learning,lecun1989backpropagation}, achieving impressive performance on a variety of computer vision tasks in recent years \cite{overfeat,rcnn,hed}. Noteworthy CNN-based detection schemes are the OverFeat \cite{overfeat} and Region-based CNN (R-CNN \cite{rcnn}) architectures. Although both employ a CNN, OverFeat performs sliding-window detection (which is common in traditional object detection), while R-CNN operates on a set of region proposals. We note that both \cite{overfeat,rcnn} operate in a local-region manner without joint reasoning over multiple scales of an image pyramid. Current improvements over such architectures emphasize 1) The learning and incorporation of deeper networks \cite{vggcnn,vgg}, 2) Resolving different components of the successful R-CNN framework into a single, end-to-end architecture. The original R-CNN framework involves a multi-stage pipeline, from object proposal generation (e.g. Selective Search \cite{selsearch}) to SVM training and bounding box regression. At test-time, a CNN forward pass is performed for each region proposal, which is costly. In contrast, SPPnet \cite{sppnet}, Fast R-CNN \cite{girshick15fastrcnn}, and OverFeat require only a single forward pass. Fast R-CNN \cite{girshick15fastrcnn} employs a Region of Interest (ROI) pooling layer which operates on region proposals projected to the convolutional feature map. Furthermore, the bounding box regression module is also integrated into the end-to-end training using a sibling output layer. Recently, another boost in performance was introduced in Faster R-CNN \cite{renNIPS15fasterrcnn}, which incorporates a Region Proposal Network in order to improve over the Selective Search region-proposal module. Independent testing at multiple scales is shown to improve performance on the PASCAL benchmark in the aforementioned studies, yet no further analysis is shown. Larger gains from multi-scale analysis are generally shown for other domains requiring robustness over large scale variations such as on-road vehicle detection \cite{Geiger2013IJRR} and pedestrian detection on the Caltech benchmark \cite{ccf,cvpr13peds}. In general, common CNN and hand-crafted object detectors involve training for and classifying a local region with a single-scale model. The contextual modeling capacity of such models is therefore limited, and detection of objects at multiple scales is done by independent scoring of an image pyramid. Nonetheless, visual information across scales at a given image location is highly correlated. Therefore, pooling features over scales in training and testing may benefit an object detector. Our work leverages a novel multi-scale detection framework in order to study the role of contextual information across image scales in a given spatial location.   

\textbf{Contextual object detection}: Our study is relevant to the study of context. Classifying scale volumes directly benefits from contextual cues found at different levels of an image pyramid. Hence, scale and context modeling are interleaved fundamental tasks in computer vision \cite{Osaku201560,nmslayout,hed,mscnn,Hoai20141523}. Careful reasoning over these two tasks has shown great success in a variety of computer vision domains, from image segmentation \cite{yannmssseg} to edge detection \cite{hed}. The Deformable Part Model (DPM) \cite{dpmPAMI,dpmbird} is another example, as it reasons over a lower resolution root and higher resolution parts templates. Commonly, an additional module for capturing spatial and scale contextual interactions is applied over the score pyramid output of a traditional local-region, single-scale detector \cite{rcnn,regionletssgm,dpmPAMI,context1}. In contrast, the studied framework in this work joins the two steps. In Chen \etal \cite{moco}, a Multi-Order Contextual co-Occurrence (MOCO) framework was proposed, extending the Auto-Context idea \cite{tuAuto,fixedpt} for context modeling among boxes produced by traditional local region detection schemes. Sadeghi and Farhadi \cite{visualphrase} propose visual phrases to reason over the output of object detectors and local context of object relationships. Desai \etal \cite{nmslayout} formulate multi-class object recognition as a structured prediction task, rescoring object boxes and replacing NMS for improved modeling of spatial co-occurrence. Li \etal \cite{li2014integrating} propose a hierarchical And-Or model for modeling context, parts, and spatial arrangements, and show large detection performance gains at a car detection task. Unlike the aforementioned, this work aims to study the benefit of incorporation of contextual, multi-scale cues directly into to object detection scheme. This is done both by modifying the detector to operate on scale volumes spanning the entire image pyramid and the inference label space. Analysis regarding the impact of such a framework is lacking in the aforementioned studies.
%A similar approach has been applied in \cite{facestruct} for face detection. 

\textbf{Multi-scale deep networks for contextual reasoning}: Multi-scale deep networks have been previously studied in \cite{mscnn,yannmssseg,sermanet2011traffic}. Eigen \etal \cite{mscnn} predicts depth maps by employing two deep network stacks, one for making coarse global prediction over the entire image and another for local refinement. Similarly to \cite{mscnn}, this work aims to analyze the role of capturing information at different image scales. In contrast to \cite{mscnn}, we discus the task of object detection and localization, study deep features at more than two image scales, and aim to better capture image appearance variations due to scale. Sermanet \etal \cite{sermanet2011traffic} propose a multi-scale branched CNN for traffic sign recognition. Here, scale refers to different levels of feature abstraction as opposed to image pyramid scales. Although related to our study in capturing context, the method does not employ feature responses or weight learning across image scales for handling scale variation and improved object localization. 

A close approach to ours is the work of Farabet \etal \cite{yannmssseg}, which proposes a multi-scale CNN for semantic scene labeling of pixels. Consequently, segmentation quality is significantly improved by learning CNN weights which are shared across three image scales. Commonly, multi-scale architectures employ 2-3 image scales at most, while we employ 7-10, and modify the inference label space. The multi-scale CNN is shown in \cite{yannmssseg} to be better at capturing image evidence at a certain pixel location, yet no insight is given regarding th impact at different object scales (e.g. small objects), contribution of weights at different scales, relationship between object class and context usefulness, or impact on localization quality. Generally, adding responses at multiple image scales is known to benefit a variety of vision tasks, yet analysis on its role for general object detection and localization is lacking. Our study is also motivated by the fact that most current state-of-the-art object detectors do not employ multi-scale features or modeling \cite{overfeat,rcnn,girshick15fastrcnn,renNIPS15fasterrcnn}. Furthermore, the training formulation in this work allows for visualization of the multi-scale, contextual cues. In contrast, most related studies discuss improvement due to multi-scale image features on a performance level only (e.g. features with one image scale vs. two image scales), without providing further insights.

\section{Capturing Context with the Multi-Scale Structure (MSS) Approach}
\label{sec:ssapproach}

The main approach in which context in object detection will be studied is presented in this section. The method is contrasted with existing schemes which are limited in their contextual reasoning in Fig. \ref{fig:main}. Instead of training and testing over local image regions (either a sliding window or region proposals), the approach employs an image pyramid and operates on features at all scales in training and testing. As large scales include fine-grained information, such as part-level information, and small scales include scene-level information, the MSS approach allows a study of the importance of cues at different scales. Furthermore, scale-specific multi-scale models are trained as contextual cues vary greatly w.r.t. the object scale. The MSS approach is also directly comparable to traditional single-scale training/testing baseline as the feature pyramid input to both is kept the unchanged.

\subsection{Efficient Feature Pyramids}
\label{sec:efficientnet}

In order to efficiently train and test models which reason and pool over multi-scale features, all experiments are performed in an architecture similar to OverFeat \cite{overfeat,slidingCNN} and DeepPyramid DPM \cite{dpmarecnn}. These have shown powerful generalization and flexibility to a variety of tasks, even without fine-tuning \cite{ccf}. Hence, they are suitable for studying the ability to model context when transferring from the ImageNet classification task to the detection task. Furthermore, they provide \textit{simple and efficient} means for handling multi-scale image pyramid information (order of magnitude faster than the original and widely used R-CNN \cite{rcnn}). By only employing the convolutional layers (discarding the fully connected layers), spatial structure is preserved and image regions can be directly projected to feature responses in an efficient manner without requiring a region proposal mechanism. Although more intricate approaches exist which preserve the fully connected layers (such as faster R-CNN \cite{renNIPS15fasterrcnn}), the used ROI pooling layer in existing approaches still \textit{operates on a single scale} of image features, and so the approach is orthogonal to our study. The network we employ is a truncated version of the winning network of the ILSVRC-2012 ImageNet challenge \cite{Krizhevsky} composed of 8 layers in total. The network is used as a main tool to better understand context in CNNs. Employing deeper networks \cite{renNIPS15fasterrcnn,vgg} greatly improves performance by improving \textit{local classification} power, but these are generally evaluated in a single-scale manner (or independent evaluation over multiple scales) and so are also orthogonal to this study. As tasks with large scale variation (e.g. pedestrian detection \cite{ccf,DollarPAMI14pyramids}) require a large image pyramid in order to reach state-of-the-art performance, the approach in this work is also motivated by the need of real-world applications for a trade-off between performance, computational efficiency, and memory requirements. Our study of efficient multi-scale contextual reasoning is directly applicable to such applications. 

\begin{figure*}[!t]
\centering
\begin{tabular}{c}
\includegraphics[trim = 0mm 85mm 0mm 0mm, clip,width=6.6in]{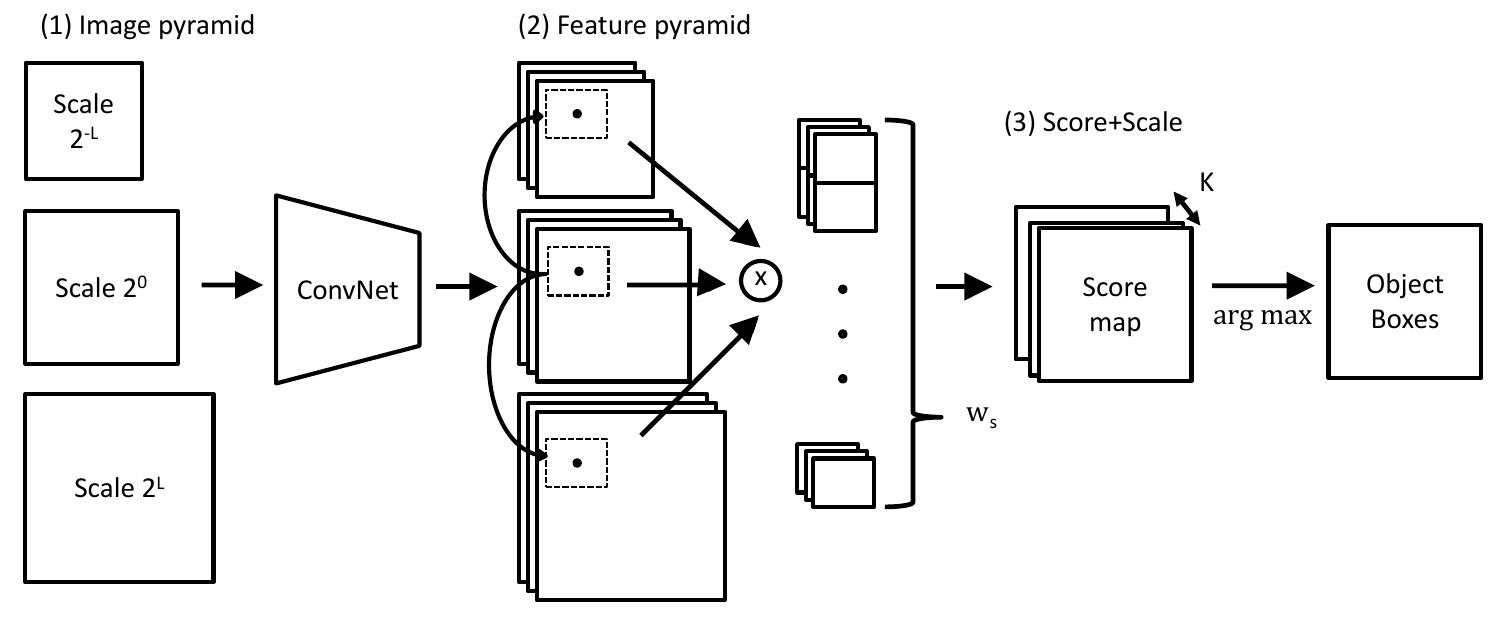} %{figs/diagramlarge.pdf}
\\
  \end{tabular}
   \caption{Our proposed approach re-samples the original image to obtain an image pyramid. Object-level annotations are converted to multi-scale annotations by obtaining a scale label. The scale label is assigned for each sample based on an overlap of the ground truth in each scale with a fixed model size (Section \ref{sec:labelspace}). Each sample is associated with a feature array that is cropped from the feature pyramid at shifted versions for preserving the same spatial location across scale. Testing involves scoring (represented by the `X' operation in the figure) using learned multi-scale templates which convert the feature pyramid to an object score map. Note that the feature maps for each scale shown in the figure are at a lower spatial resolution than the original images.}
 \label{fig:diagram}
\end{figure*}   

\subsection{Multi-scale detection with a single-scale template}

First, we introduce notation to clarify and motivate the MSS approach. In traditional object detection, context reasoning is limited as detection is performed in a single scale fashion (tested independently at multiple image scales). First, a feature pyramid is constructed over the entire image at each scale to avoid redundant computation for each striding window. Let $p_s=(x,y,s)$ be a window in the $s$-th level of a feature pyramid with $S$ scales anchored in the $x,y$ position. Most of the analysis will involve a single aspect ratio model (which is common), and so we do not include that additional parameter in $p_s$, yet the formulation supports multiple aspect ratio models \cite{subcat}. Generally, the feature pyramid is at a lower spatial resolution than that of the image of the same scale (due to convolution and sub-sampling). Consequently, a zero-based index $(x,y)$ in the feature map can be mapped to a pixel in the original image using a scale factor $(cx,cy)$ based on the resolution of the feature map. Mapping locations over scales can be achieved by a multiplication by the scale factor as well.
 Each window contains an array of feature values, $\phi(p_s) \in \mathbb{R}^d$, to be scored using a filter $w$ learned by a discriminative classifier, in our case a support vector machine (SVM). The scoring is done using a dot product,

\begin{equation} 
\label{eqn:ipyr}
	{f}(p_s) = w \cdot \phi(p_s)
\end{equation}

  Generally, the template size is defined as the smallest object size to be detected, and further reduction in template size results in degradation of the detection performance. Note that learning and classification only occurs over a local window. A similar pipeline can be described using a template pyramid as studied in \cite{peds100rodrigo,30hzdpm,parkmultires} and was shown to improve results due to capturing finer features at different scales that would have been discarded by the down-sampling. In this approach, a set of templates are learned, $(w_1,\dots,w_{S})$. In detection, the $S$ templates are evaluated so that each location $p$ in the original image scale is scored using the set of model templates
  
  \begin{equation} 
  \label{eqn:tpyr}
	{f}(p) = \max_{s\in\{1,\dots,S\}} w_s \cdot \phi(p)
\end{equation}
  
  where we drop $s$ as only one scale of the image is considered. We emphasize that the model filters in this approach are also trained on locally windowed features only, but may capture different cues for each scale. In principle, this approach is similar to the baseline as it performs the scoring convolution at each scale independently of all other scales (unlike MSS, as shown in Fig. \ref{fig:diagram}).

\subsection{Multi-scale detection with a multi-scale template}
\label{sec:MSS}

The feature pyramid computation and handling is mostly left unchanged in the proposed MSS approach. Spatial locations in the image space can be mapped across scales using a scale factor. As shown in Fig. \ref{fig:motiv1}, evaluations at the same spatial location occur repeatedly over scales. This mechanism is replaced by considering features from all scales at a given image location, i.e. $\psi(p) = (\phi(p_1),\dots,\phi(p_S)) \in \mathbb{R}^{d \times S}$ descriptor.  

\subsubsection{Label space}
\label{sec:labelspace}

Next the process of labeling training samples is outlined. Each sample is assigned a label, $y = ({y}^l,{y}^b,y^s)\in \mathcal{Y}$ with $y^l$ the object class (in this study only ${y}^l \in \{-1,1\}$ is considered), $y^b \in \mathbb{R}^4$ is the object bounding box parameters, and $y^s$ is a scale label. In our experiments, the model dimensions are obtained from the average box size of all positive instances in the dataset (providing a single aspect ratio model). Training instances are sampled directly from the feature pyramid in a simple process where, 1) the multi-scale template is centered on top of each ground truth window spatial location and 2) Overlap with the ground truth is checked in each image scale (as shown in red in Fig. \ref{fig:motiv1}). Formally, a vector of overlaps $F$ is constructed. If the image at $s$-th level contains $\hat{y}(s) = \{\hat{y}_1(s),\dots,\hat{y}_N(s)\}$ ground truth boxes, the template box is centered on a positive sample at the $s$-th level (denoted as $B(s)$), so that entries of $F$ are computed for each pyramid level,

\begin{equation}
 \label{eqn:scalevector}
    F(s) = \max_{i\in \{1,\dots,N\}} \text{ov}(B(s),\hat{y}_i^b(s)).
  \end{equation} 
  
  %Changed to (s) to not confuse with scale label y^s 

  where ov($a$,$b$) = area($a\cap b$)/area($a \cup b$) for two rectangles, $a$ and $b$. $F$ is shown for three examples in Fig. \ref{fig:motiv1}. For instance, for Fig. \ref{fig:motiv1} first row, $y^s = (0100000)$. Peaks in $F(s)$ with high overlap imply a positive instance. This process potentially allows for multiple labels over scales to be predicted jointly, i.e. two almost overlapping objects at different scales, but such instances are rare. For simplicity, we only allow a single scale-label association by employing the scale where maximum overlap occurs.

\subsubsection{Learning}
	
	Two max-margin approaches are studied for learning the multi-scale object templates, leveraging the highly structured multi-scale information, and analyzing importance of contextual information at different scales. Such information would have been ignored if a single-scale template was used. 
	
Parameterization in the image pyramid can be done once over spatial locations at different scales by mapping across region locations with a scaling factor. Although these local regions across scales remain the same both in a traditional single-scale model classification procedure and the MSS approach, this new parameterization implies that we can concatenate features at all scales, as opposed to classifying these separately across scales. Furthermore, the previous section showed how such samples could be labeled, so the problem can now be posed as a multi-class problem. 

\textbf{One-vs-All}: There are well developed machine learning tools for dealing with a large-dimensional multi-class classification problem. A straightforward solution is with a one-vs-all (OVA) SVM, which allows training the multi-class templates quickly and in parallel. 
Window scoring is done using

  \begin{equation} 
  \label{eqn:ova}
	{f}(p) = \max_{s\in\{1,\dots, K\}} w_s \cdot \psi(p)
\end{equation} 
The scale of the box is obtained with an $\argmax$ in Eqn. \ref{eqn:ova}.
In order to learn the $K$ linear classifiers parameterized by the weight vectors $w_s \in \mathbb{R}^{d \times S}$, the stochastic dual coordinate ascent solver of \cite{vedaldi08vlfeat} with a hinge loss is used. The maximum number of iterations is fixed at $5 \times 10^6$ and the tolerance for the stopping criterion at $1 \times 10^{-7}$ for all of the experiments. Training a single multi-scale template on a CPU on average takes less than a minute. For simplicity, this paper considers training a model for each scale, so that $K = S$. In general, this may not be the case (e.g. pedestrians occurring at close proximity but at different scales). 
 
\textbf{Structured SVM}: A second approach can be used in order to learn all of the multi-scale templates jointly. A feature map is constructed using the labels of each sample as following,

\begin{equation}
\Phi(p,y) = (\Psi_1(p,y),\dots,\Psi_K(p,y)).
\end{equation}

 \begin{equation}
   \Psi_k(p,y)=
    \begin{cases}
      \psi(p) & \text{if } y=k
      \\
      0 & \text{otherwise}
    \end{cases}
  \end{equation}

This approach allows for learning a joint weight vector over all classes $w = (w_1,\dots,w_{K})$, such that

  \begin{equation} 
  \label{eqn:ssvm}
	{f}(p) = \max_{y \in \mathcal{Y}} w \cdot \Phi(p,y) 
\end{equation} 

Where the scale label prediction similar to as in Eqn. \ref{eqn:ova}, but the loss function in training is defined differently using other elements of $y$.    

Given a set of image-label pairs of the form $\{p^i,y_i\}$, the model is trained using a cost-sensitive SVM objective function \cite{joachims,branson,LacosteJulien2013}
\begin{equation}
\begin{aligned}
& \min_{{w},{\xi}} \frac{1}{2} || {w} || ^2 + C \sum_{i=1}^n \xi _i \\
& \text{s.t. for } \forall i, \bar{y} \in \mathcal{Y} \setminus y_i & \\ 
&   {w} \cdot (\Phi(p^i,y_i) - \Phi(p^i,\bar{y})) \geq L(y_i,\bar{y})-\xi _i 
\end{aligned}
\end{equation}

The loss function, $L$, is chosen to favor large overlap with the ground truth,

 \begin{equation}
    L(y,\hat{y})=
    \begin{cases}
      0 & \text{if }
       y^l = \hat{y}^l = -1 \text{ or} \\
       & \displaystyle\max_{i\in \{1,\dots,N\}} \text{ov}(y^b,\hat{y}_i^b) < 0.6
       \\
      1 & \text{otherwise}
    \end{cases}
  \end{equation}
  %\text{ov}(y_i^b,y^b) < 0.6
  
% The training loss function is then taken as an average loss over each of the predicted boxes, $\bar{L} =\frac{1}{M} \sum_{j\in\{1,\dots,M\}} L(y_j,\hat{y})$, for $M$ predicted boxes.
  
 \subsubsection{Generalization of the single-scale approach}

The main aim is to study context. The purpose of introducing the MSS approach is that it generalizes the traditional single-scale approach. Below, we show that in principle, if other scales do not contain additional contextual information, MSS reduces to the traditional single-scale approach. Eqns. \ref{eqn:ova} and \ref{eqn:ssvm} employ features at all scales for a given spatial location. Such a formulation allows learning the class weights jointly, as in Eqn. \ref{eqn:ssvm}. It can be shown that this is a generalization of the single-scale template baseline. For instance, if no discriminative value is added by adding features at different scales, then the corresponding weights $w_s$ in Eqn. \ref{eqn:ova} will only select features in the single best-fit scale (i.e. a degenerate case). Therefore, for each level $s$ in the pyramid, $w_s \cdot \psi(p)$ becomes identical to $w \cdot \phi(p_s)$ as in Eqn. \ref{eqn:ipyr}. A similar argument demonstrates the same for Eqn. \ref{eqn:ssvm}. Therefore, both of the studied multi-scale template learning approaches can benefit by having access to additional information not accessible to the single-scale template approaches which only employs local window features at one scale. Furthermore, by learning a separate weight for each class, the model can account for appearance variations at different resolutions \cite{peds100rodrigo} and learn scale-specific context cues.

\section{Experimental Evaluation}

The experiments aim to quantify the importance of context cues in deeply learned features for a detection and localization task. Initially, the MSS approach is developed on the PASCAL VOC 2007 dataset \cite{pascalvoc} using its established metrics, followed by analysis on a multi-view highway vehicles dataset with large variation in object scale.

\begin{figure}[!t]
\centering
\begin{tabular}{c}
  \includegraphics[width=2in]{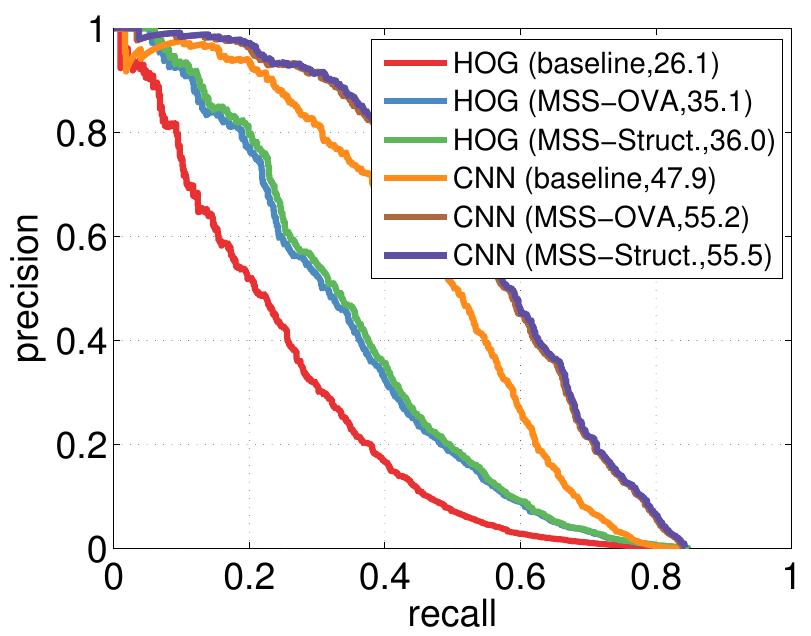}
 \end{tabular}
 \caption{Model training comparison on a validation set for `car' detection using HOG and conv$_5$ features. Average Precision (AP) is shown in parenthesis. Contextual information captured with MSS is shown to significantly improve detection performance using both one-vs-all (OVA) and structural SVM (Struct.) training.}
 \label{fig:resMiningParams}
\end{figure}

\begin{figure*}[!t]
\centering

\resizebox{15cm}{!} {
\begin{tabular}{c}

\begin{tabular}{cc}
  Bottle-HOG & Bottle-CNN
  \\
  \begin{tabular}{c}
  %trim = 0mm 100mm 120mm 0mm, clip,
%[-0.6ex]
\fbox{\includegraphics[width=3.1in]{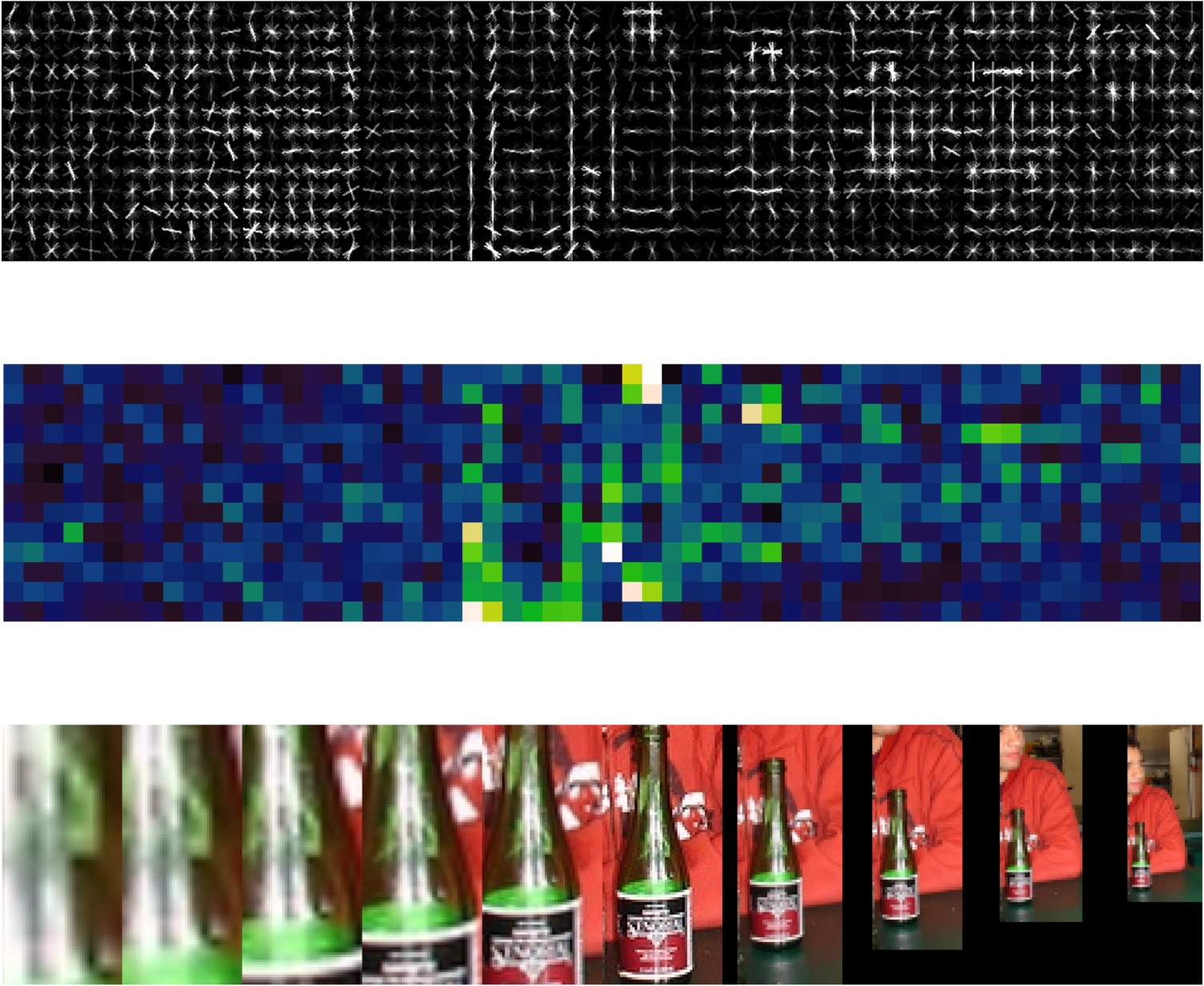}}
 \end{tabular}
 &
 \fbox{
   \begin{tabular}{c}
  % \hline
%\\
%[-2.5ex]
\includegraphics[width=1.5in]{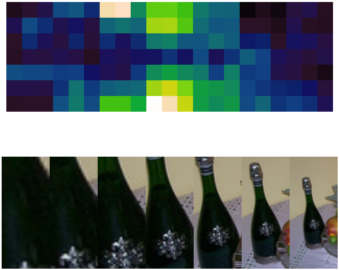}
 \\
 \\
 \includegraphics[width=1.5in]{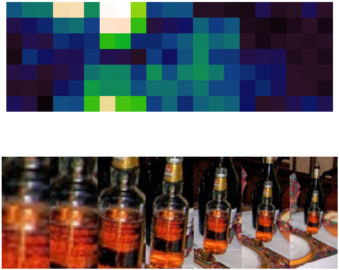}
% \\
% \hline
 \end{tabular}
 }
 \end{tabular}
\\
 \\
 \begin{tabular}{cc}
 TV monitor-HOG & TV monitor-CNN
  \\
  \begin{tabular}{c}
  %trim = 0mm 100mm 120mm 0mm, clip,
%[-0.6ex]
\fbox{\includegraphics[width=3.3in]{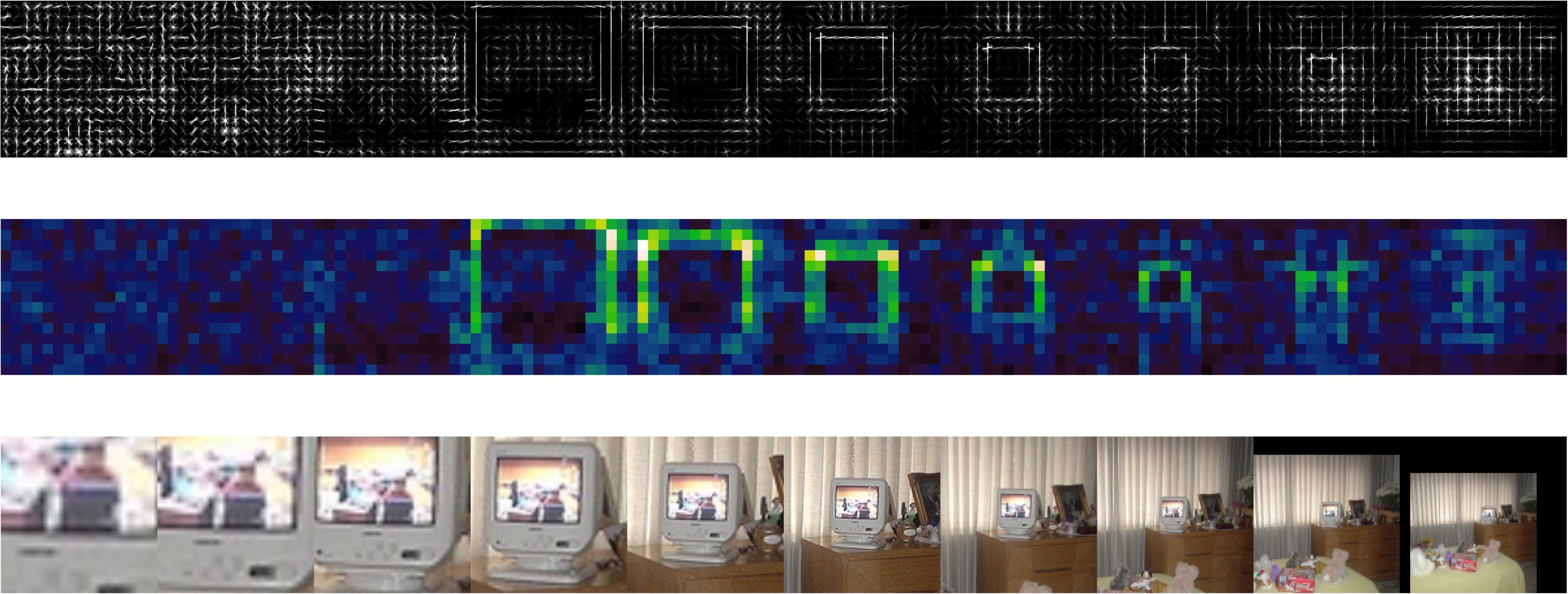}}
 \end{tabular}
 &
\fbox{   \begin{tabular}{c}

\includegraphics[width=2.1in]{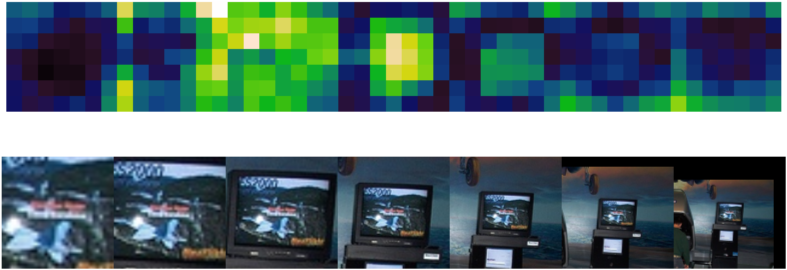}
 \\
%%% \\
 \includegraphics[width=2.1in]{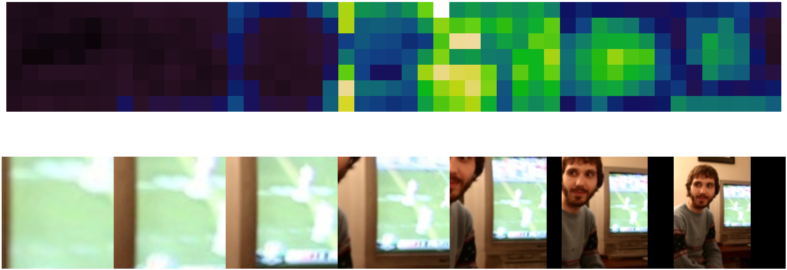}
 \end{tabular}
 }
 \end{tabular}
 \\
  \begin{tabular}{c}
 Dining Table-CNN
  \\
  \begin{tabular}{c}
  %trim = 0mm 100mm 120mm 0mm, clip,
%[-0.6ex]
\fbox{\includegraphics[width=6in]{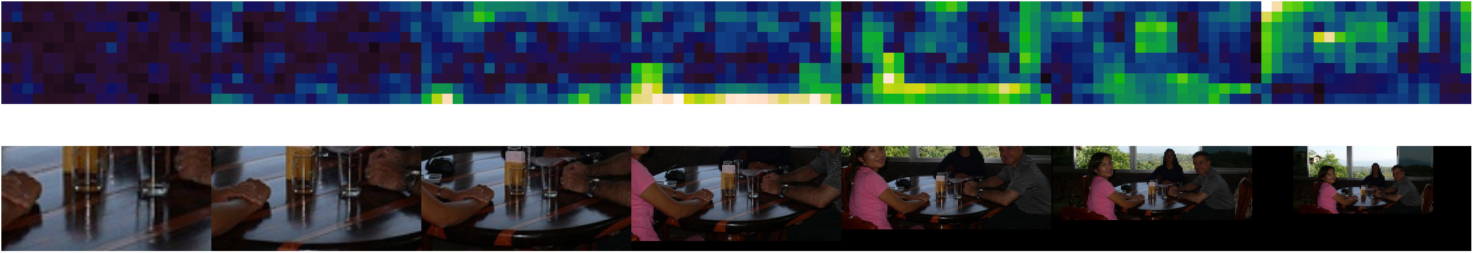}}
 \end{tabular}
\\
%%%\\
  \begin{tabular}{c}
 Aeroplane-CNN
  \\
  \begin{tabular}{c}
  %trim = 0mm 100mm 120mm 0mm, clip,
%[-0.6ex]
\fbox{\includegraphics[width=6in]{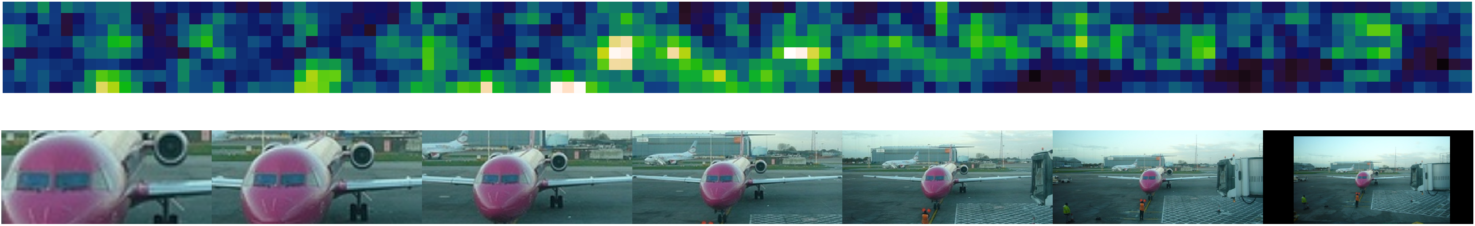}}
 \end{tabular}
 \end{tabular}
 \\
  \begin{tabular}{c}
 Car-HOG
  \\
  \begin{tabular}{c}
  %trim = 0mm 100mm 120mm 0mm, clip,
%[-0.6ex]
\fbox{\includegraphics[width=6in]{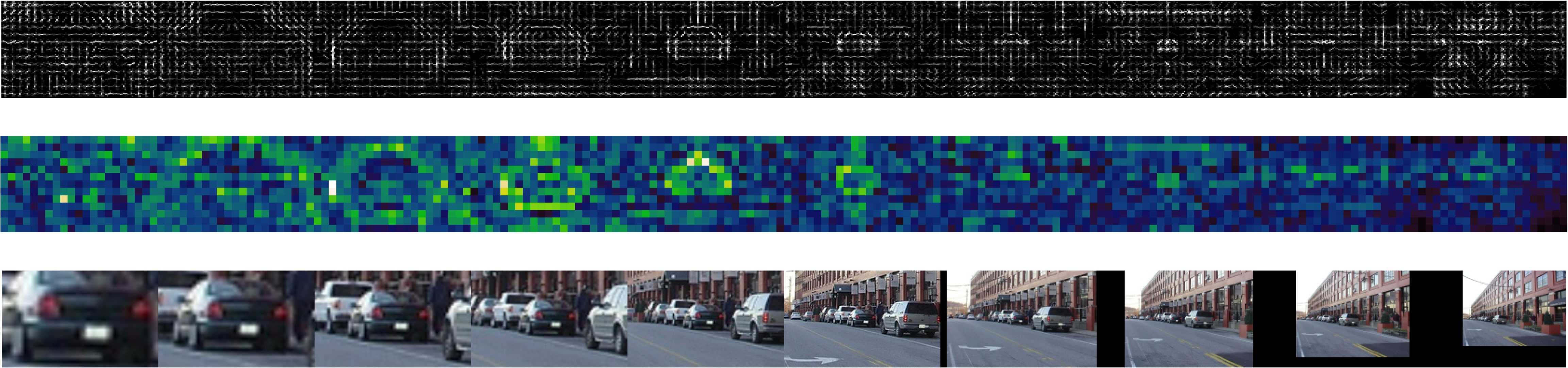}}
 \end{tabular}
 \end{tabular}
 \end{tabular}
 \end{tabular}
 }
 \caption{Visualization of multi-scale CNN and HOG templates. For each model, the maximum positive SVM weight for each block is shown together with an example instance. Brighter colors imply higher discriminative value. Large amount of discriminative value is placed at nearby and remote scales corresponding to contextual information (e.g. road cues at other scales).}
 \label{fig:vismodelspascal}
\end{figure*} 

\textbf{Features}: Two representative visual descriptors are employed in order the study the role of context. Most of the experiments involve the deeply learned features discussed in Sec \ref{sec:efficientnet}. The fifth convolution layer output has 256 feature channels. The input to each convolutional or max pooling layer is zero-padded so that the features in a zero-based pixel location $(x,y)$ in the feature space were generated by a receptive field centered at $(16x,16y)$ in the image space (a stride of 16). As noted by Girshick \etal \cite{dpmarecnn}, the CNN features already provide part and scale selective cues. This can be enhanced by applying a $3 \times 3$ max-pooling layer. For direct comparison with the DeepPyramid approach \cite{dpmarecnn}, the same feature extraction and 7-scale pyramid pipeline was implemented in the experiment. The HOG feature implementation of \cite{dpmPAMI} serves as a comparative baseline and studying generalization of experimental analysis across different feature types. HOG is used with a cell size/stride of 8.

\textbf{Image pyramid}: The scale factor between levels is set to $2^{-1/2}$. The CNN feature pyramid spans three octaves with 7 levels. For HOG features, adding 3 more levels to the image pyramid for a total of 10 was shown to improve performance. In all of the experiments, training instances are extracted directly from the feature pyramid, as opposed to extracting features from cropped image samples. For the CNN feature pyramid, the features used are computed by the fifth convolutional layer which has a large receptive field of size $163 \times 163$ pixels. 

\textbf{Data augmentation}: Training images are scale-jittered by up to an octave (either down-sampled and zero-padded or up-sampled and center-cropped). In addition to flipping, this data augmentation was essential for obtaining good performance of the MSS approach on all of the object categories. 

\textbf{Hard negative mining}: All approaches studied employ an iterative process by which hard negatives are collected for re-training. The process eventually converges, when the number of negative samples generated are below a certain threshold. All of the experiments begin with a random set of 5000 negative samples. For a given object category, the initial negative samples are kept the same across techniques to allow direct comparison. In each iteration, up to 5000 additional negatives are collected. For mining, both images containing positive instances and negative images are used. A threshold of $0.3$ overlap is used for mining negative samples from images with object instances.

\subsection{Analysis on the PASCAL VOC dataset}
\label{sec:virtual}

\textbf{Learning framework choice}: First, we evaluated the choice of learning framework on a validation set of the `car' category. Fig. \ref{fig:resMiningParams} details the analysis of different learning and features combinations on the car category. Context is shown to benefit both HOG and conv$_5$ CNN features, as both learned MSS detectors are shown to greatly outperform the baseline in detection Average Precision (AP). Training the templates using the structural SVM allows for joint learning of the MSS templates, yet the improvement is marginal. Because structural SVM training is more costly, one-vs-all models are employed for the remainder of the experiments in this study. The structural SVM formulation may be of interest in the future for bounding box regression \cite{ssvmlocalization} or parts integration \cite{dpmPAMI}.

\textbf{Visualization of the learned models}: Fig. \ref{fig:vismodelspascal} depicts some of the learned MSS models for different object categories (positive valued entries in a learned MSS weight model). A single multi-scale template is visualized with a corresponding positive instance for each object category. For a given spatial location in the model, we visualize the learned model weights at each scale. As shown, while the best-fit scale includes large amount of the discriminative value, features from other scales (both adjacent and remote) are also selected. Contextual patterns can be seen, such as selection of road cues for car detection. We also observe the existence of alignment features, where certain appearance cues at one scale may assist in localization at another scale. This is shown by a repetitive shape pattern across the scales.

\begin{figure*}[!t]
\centering
\begin{tabular}{ccc}

 \includegraphics[trim = 0mm 0mm 0mm 0mm, clip,width=2.8in]{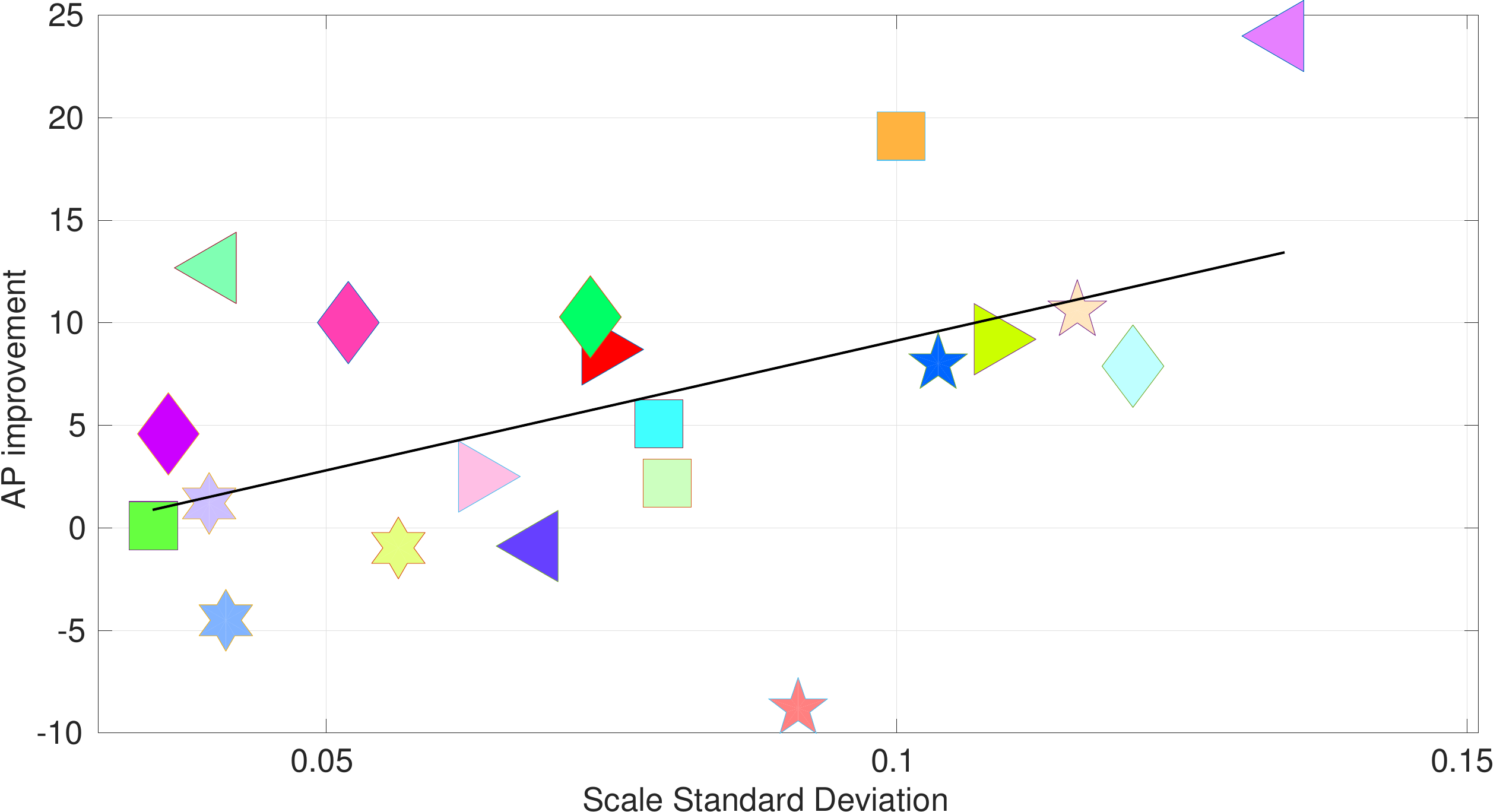}
 & \raisebox{+23ex}{\multirow{2}{*}{
 \includegraphics[trim = 153mm 0mm 190mm 0mm, clip,width=0.6in]{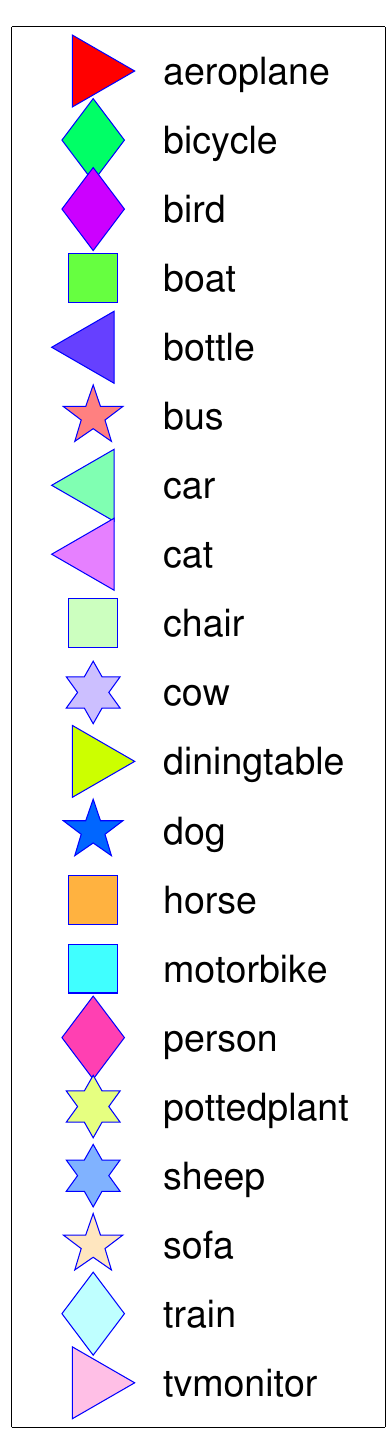}}}
&
 \includegraphics[trim = 0mm 0mm 0mm 0mm, clip,width=2.8in]{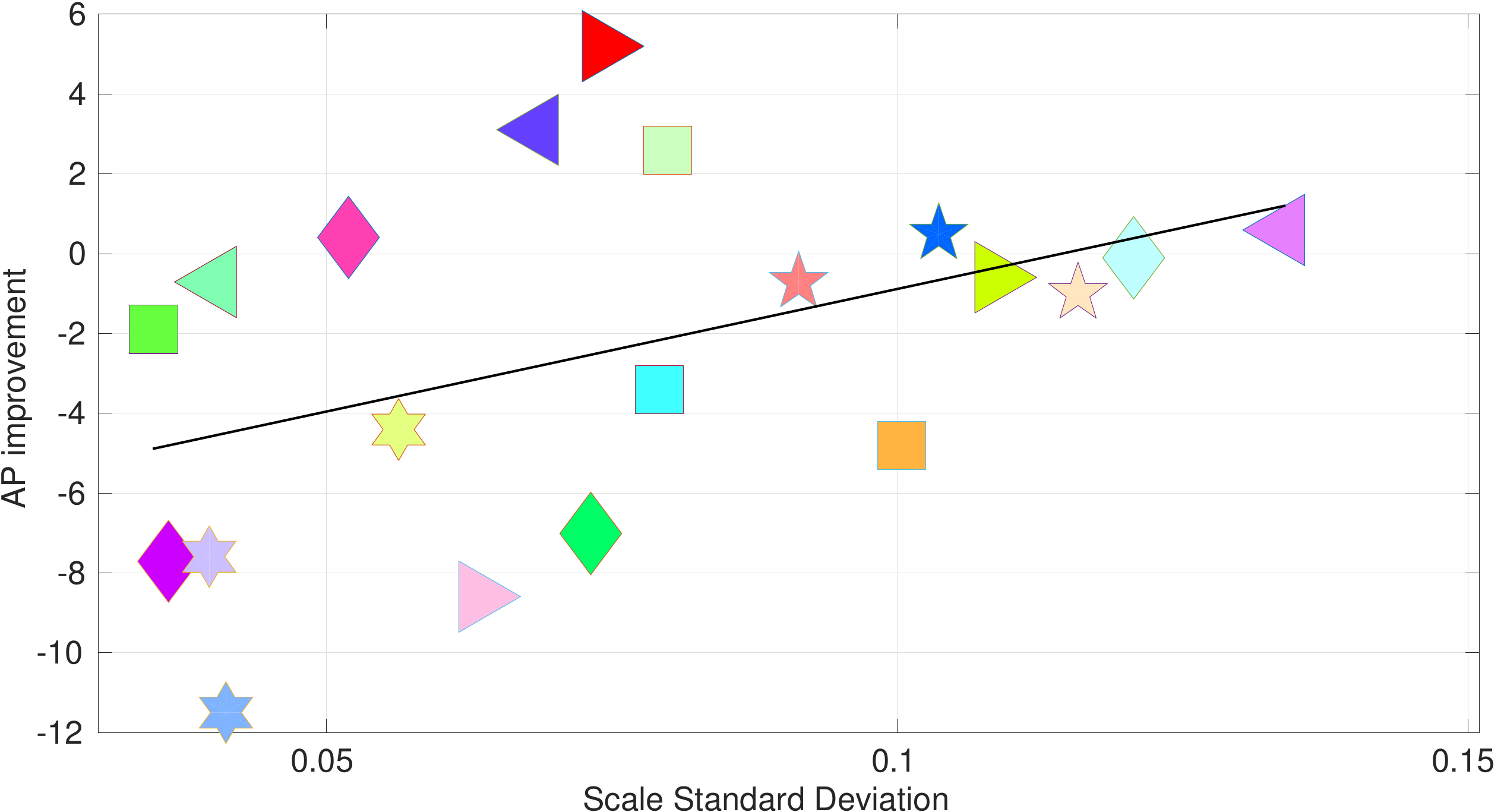} 
 \\
 (a) MSS-HOG, no data augmentation & & (b) MSS-CNN, no data augmentation
 \\
  \includegraphics[trim = 0mm 0mm 0mm 0mm, clip,width=2.8in]{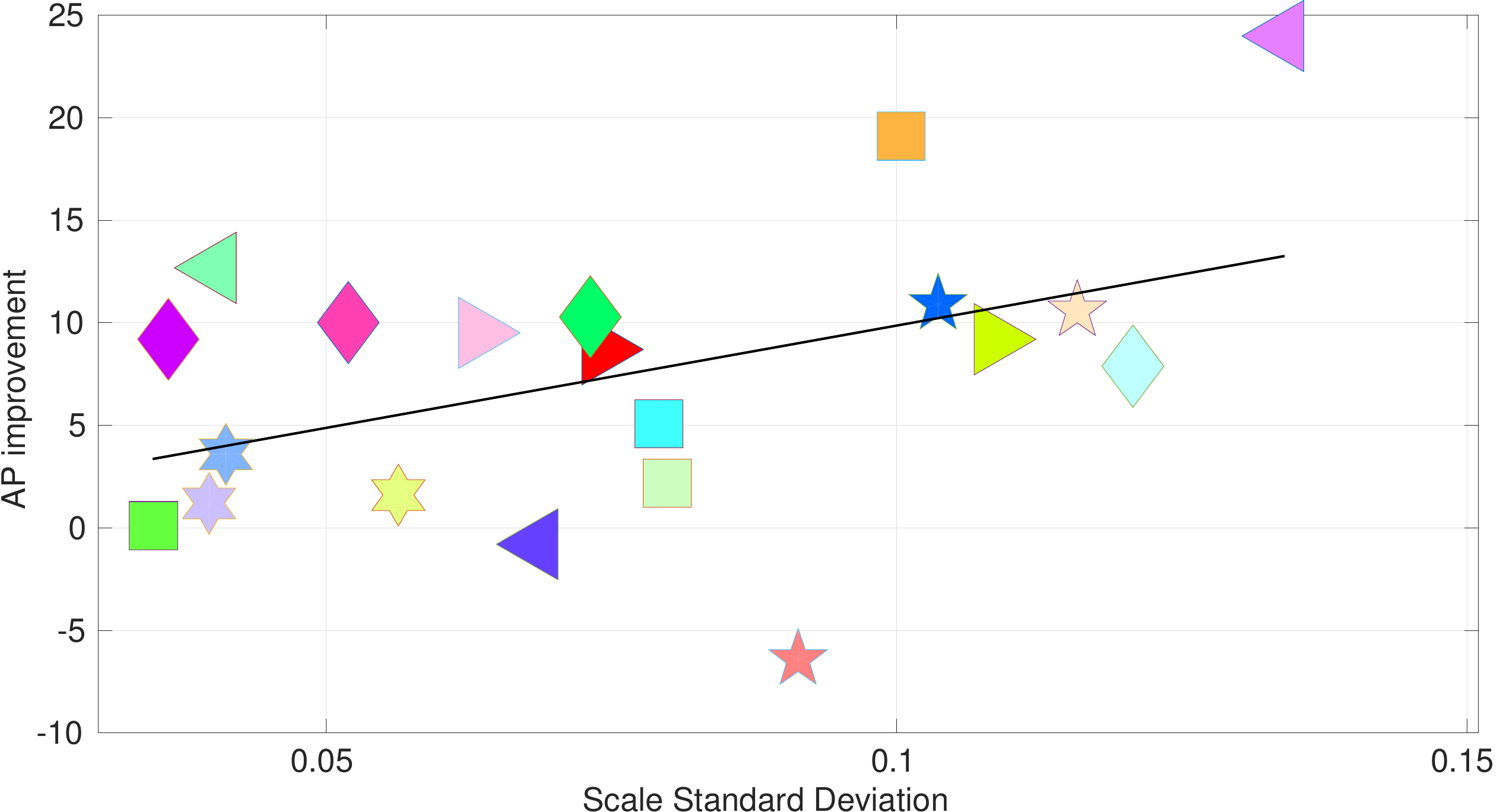}
 &
&
 \includegraphics[trim = 0mm 0mm 0mm 0mm, clip,width=2.8in]{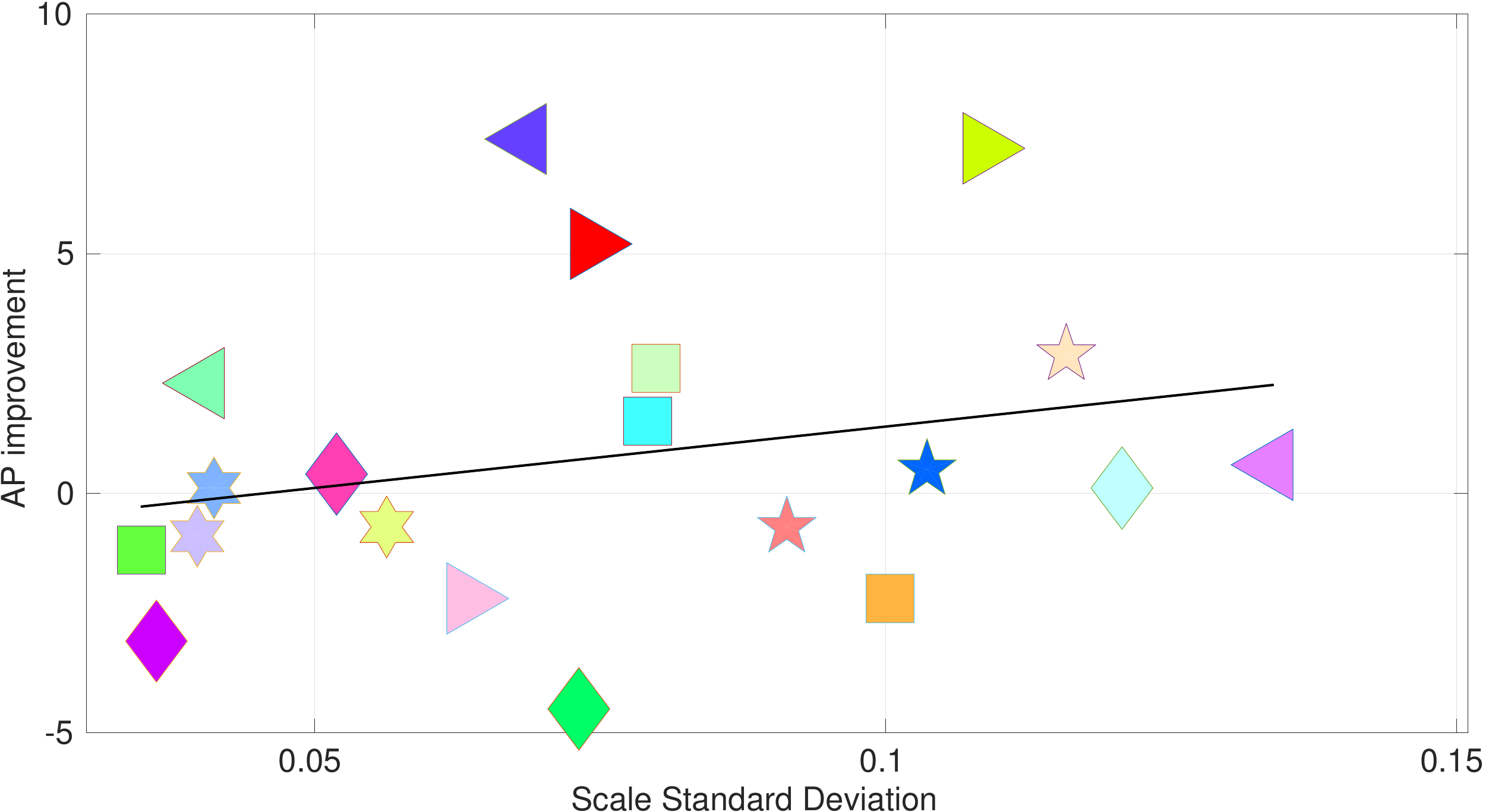} 
 \\
 (c) MSS-HOG, with data augmentation & & (d) MSS-CNN, with data augmentation

 \end{tabular}
 \caption{Relationship between the scale distribution of class samples in test time and the corresponding improvement in AP with the proposed MSS approach. As shown, our method shines when there is a large spread in the distribution over scales. Although some classes tend to appear in the PASCAL VOC dataset in a narrow scale distribution, this phenomenon is dataset and object specific. Therefore, if more instances at varying scales were to be added, the proposed approach would be better suited for such settings.}
 \label{fig:HOGscaledstbn}
\end{figure*}

\begin{figure*}[!t]
\centering
\begin{tabular}{cc}
\includegraphics[width=3in]{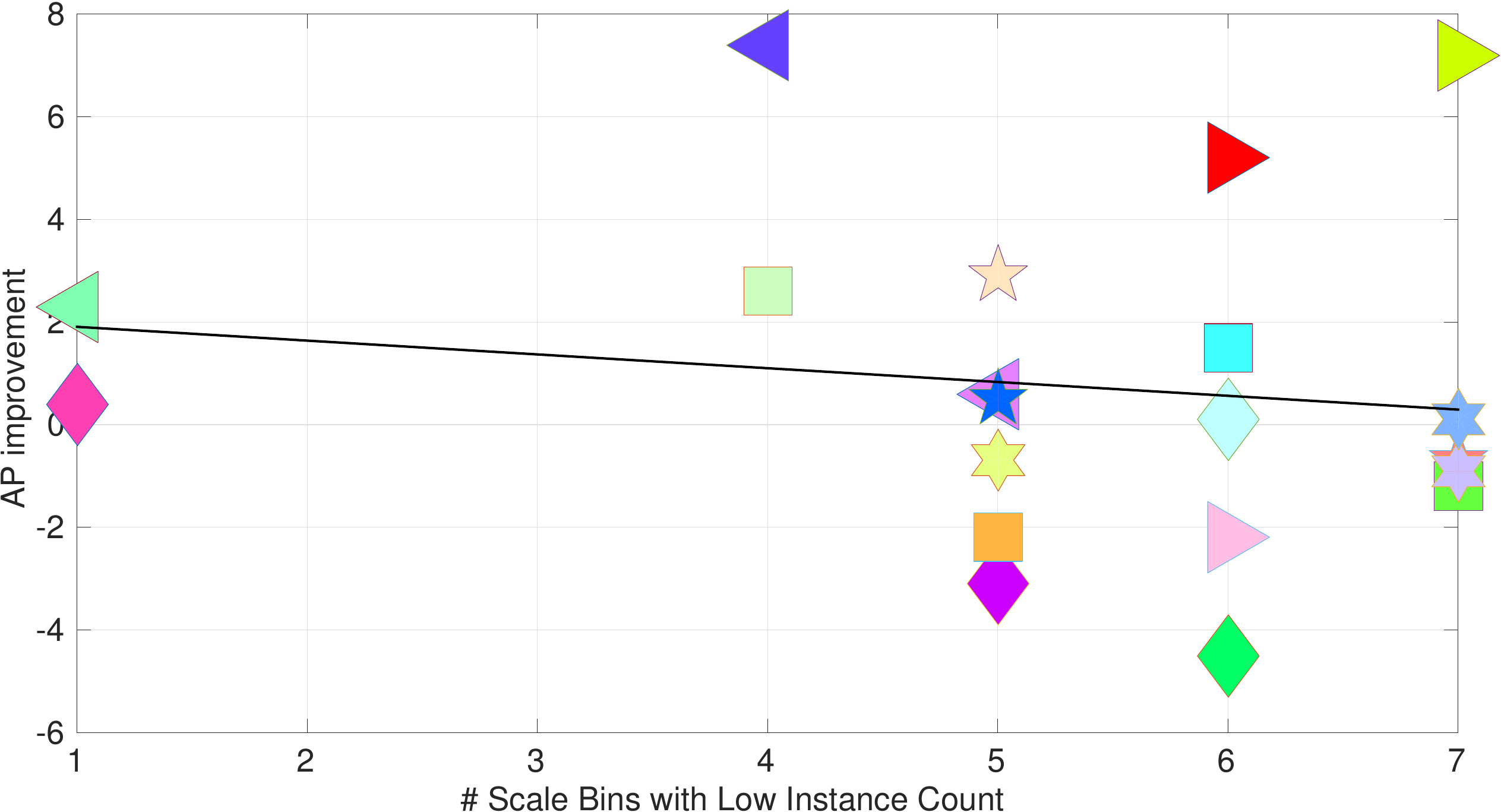}
&
\includegraphics[width=3in]{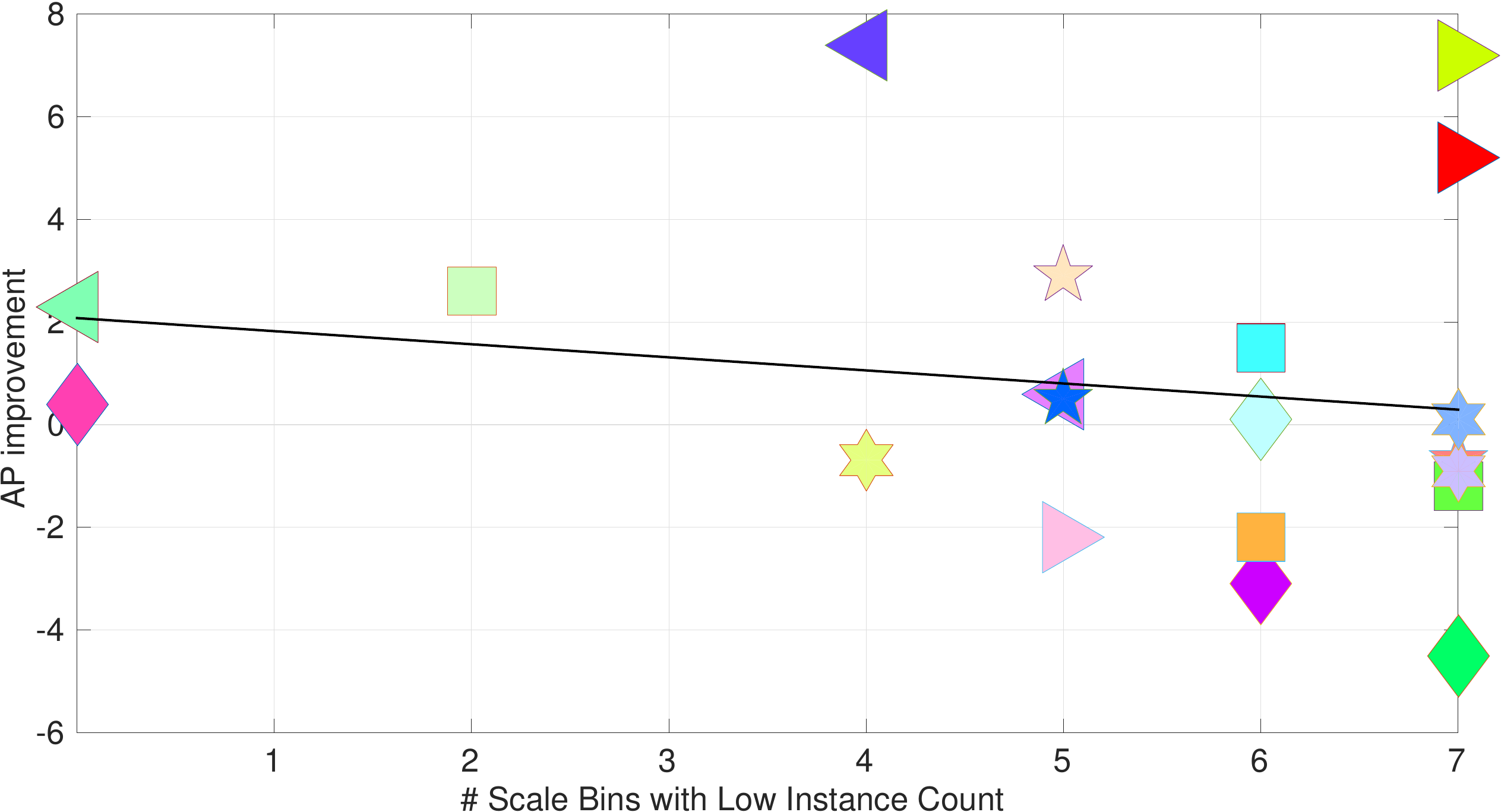}
\\
(a) Training set & (b) Testing set
 \end{tabular}
 \caption{Relationship between dataset properties and performance of the CNN-MSS approach. Some of the object classes in the PASCAL VOC benchmark contain a small number of object instances at multiple object scales, which poses a challenge to the scale-specific MSS models.}
 \label{fig:datasize_train}
\end{figure*}

\begin{figure*}[!t]
\centering
\resizebox{17cm}{!}{
\begin{tabular}{ccc}

\includegraphics[trim = 2mm 129mm 70mm 2mm, clip,width=1.8in,height=1.8in]{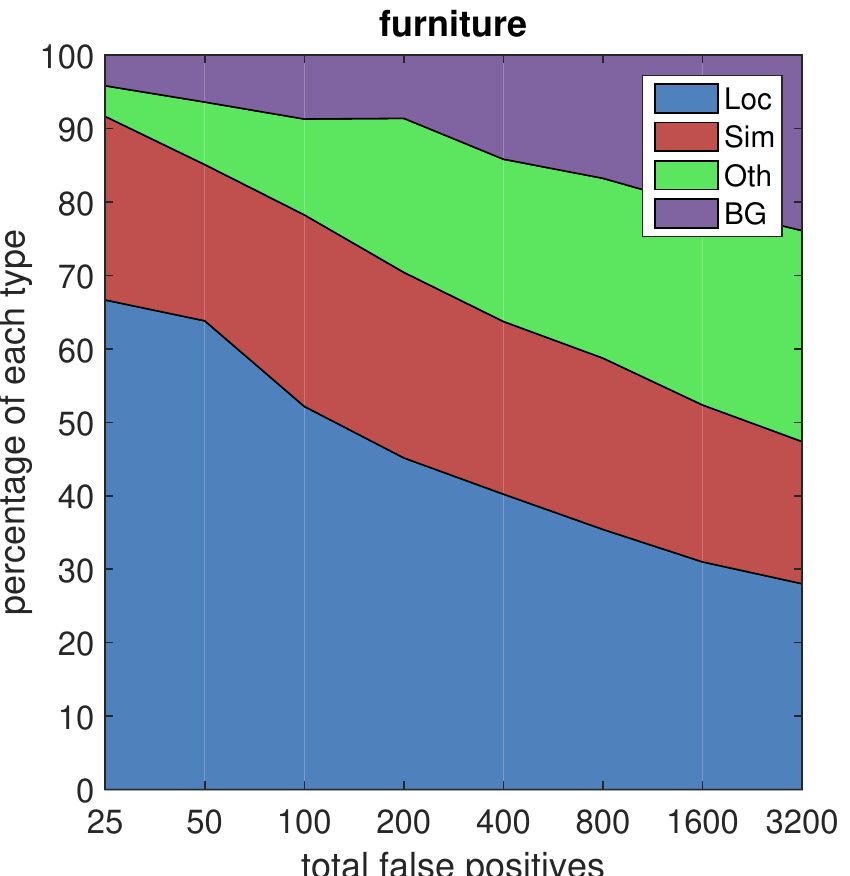}
 &
  \includegraphics[trim = 2mm 129mm 70mm 2mm, clip,width=1.8in,height=1.8in]{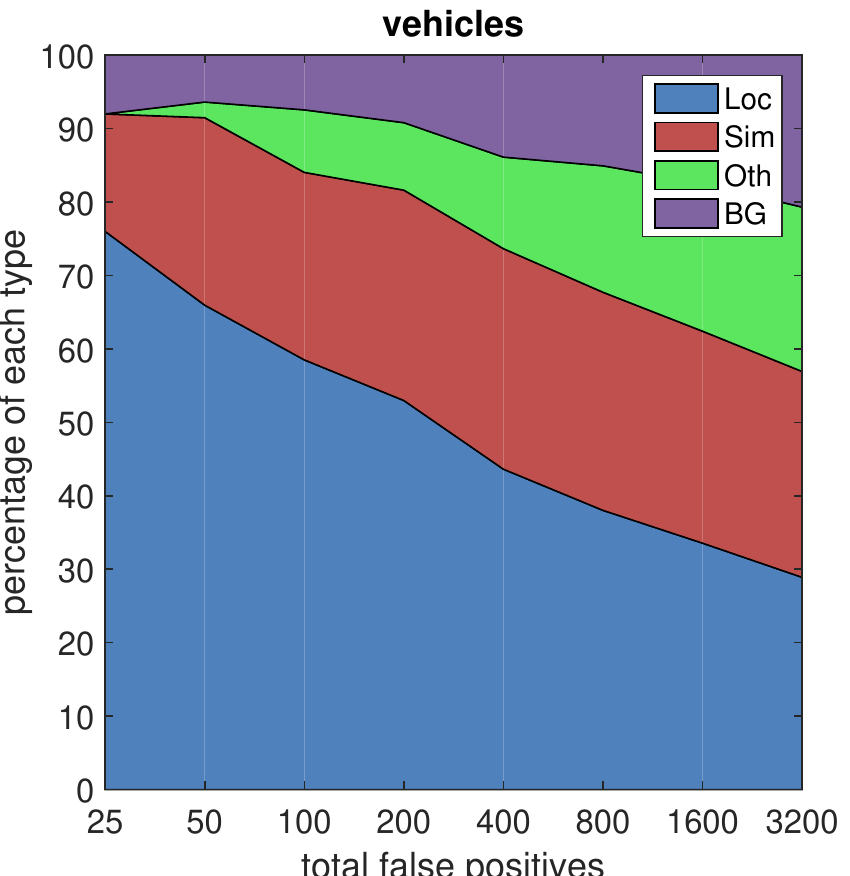}
   &
  \includegraphics[trim = 2mm 129mm 70mm 2mm, clip,width=1.8in,height=1.8in]{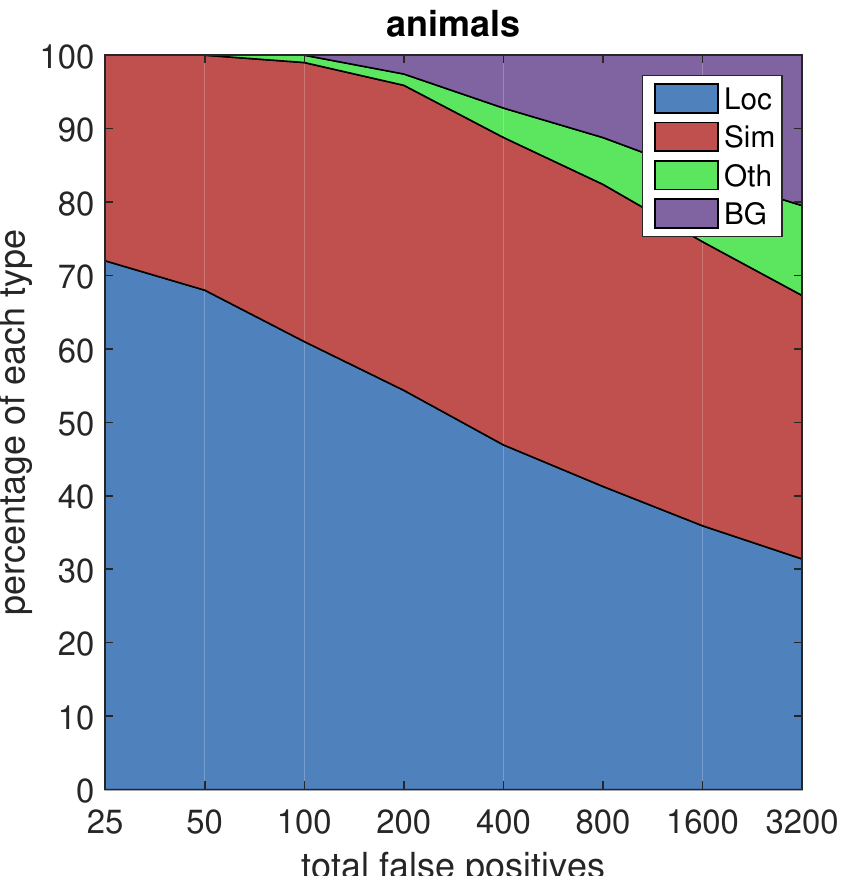}

 \\
 \multicolumn{3}{c}{(a) CNN-baseline}
 \\
  \includegraphics[trim = 2mm 129mm 70mm 2mm, clip,width=1.8in,height=1.8in]{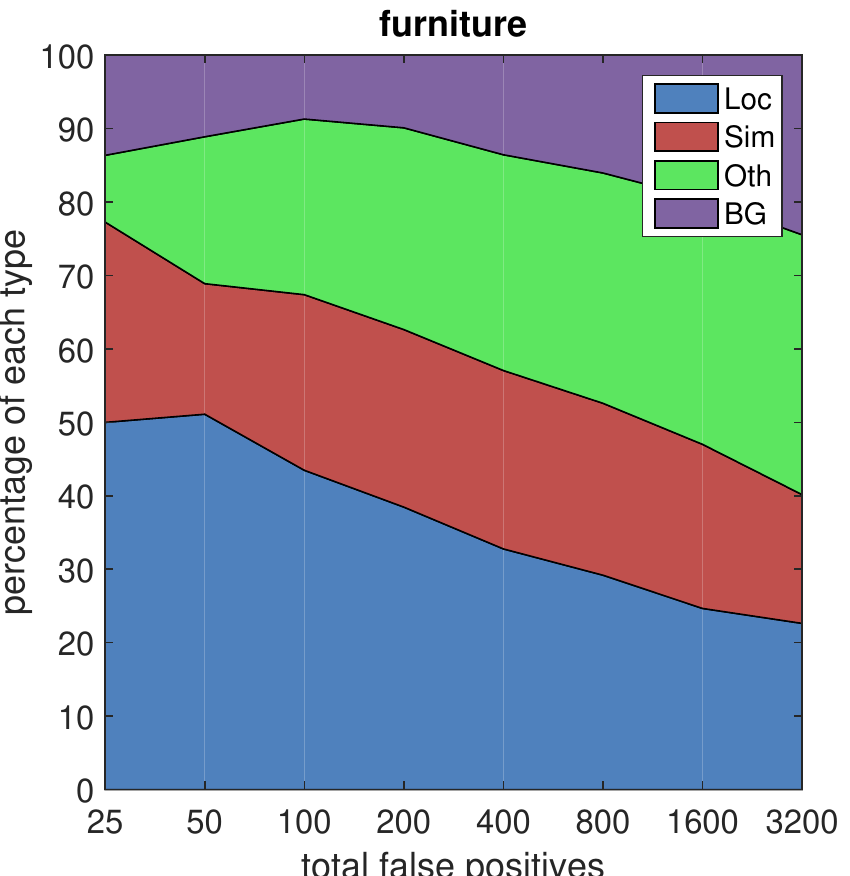}
 &
  \includegraphics[trim = 2mm 129mm 70mm 2mm, clip,width=1.8in,height=1.8in]{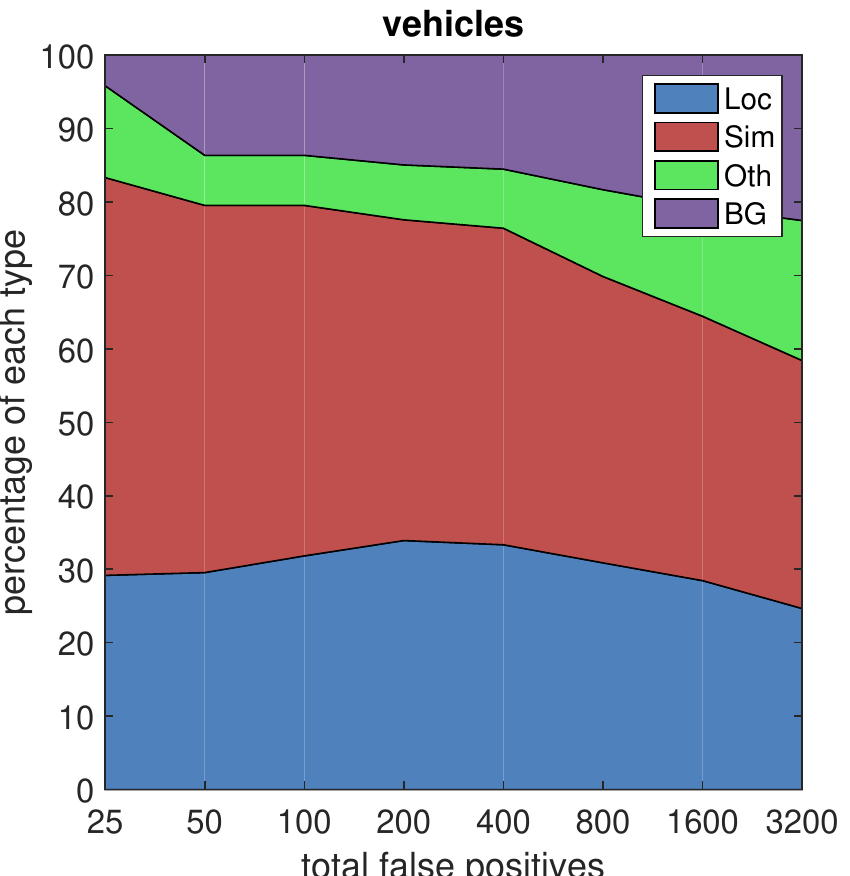}
   &
  \includegraphics[trim = 2mm 129mm 70mm 2mm, clip,width=1.8in,height=1.8in]{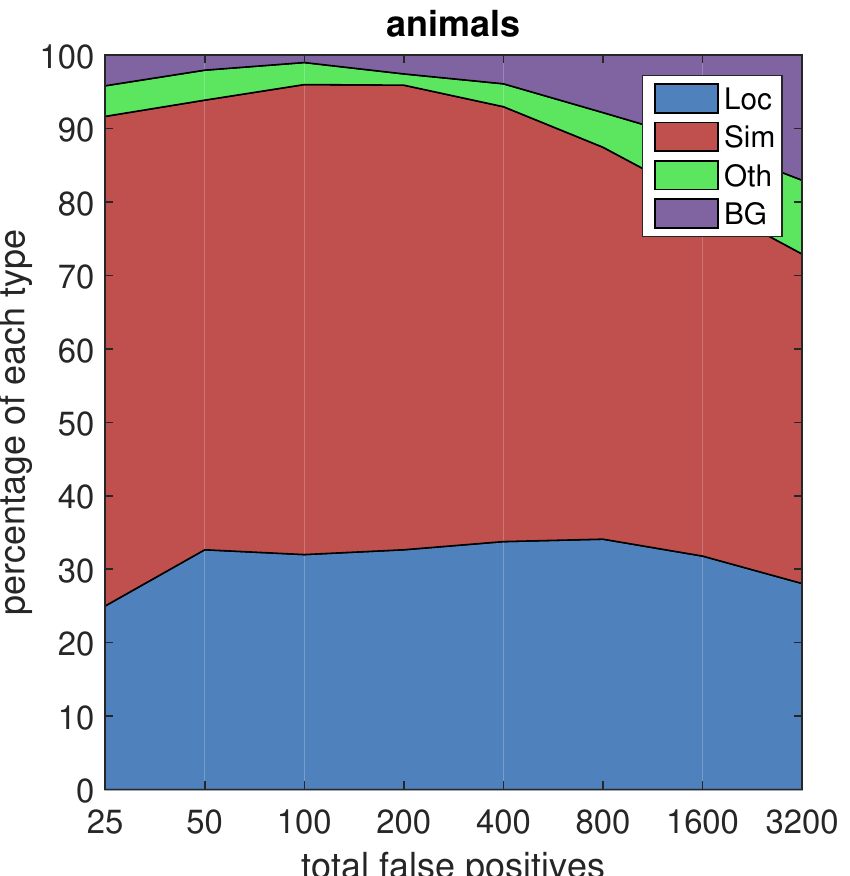}
  \\
 \multicolumn{3}{c}{(b) CNN-MSS}
 \\
  \end{tabular}
  }
 \caption{Analysis of the distribution of false positive types \cite{hoiem} for different types of objects on PASCAL VOC 2007. Training and testing is done with a single aspect ratio model. Loc - poor localization, Sim - confusion with a similar category, Oth - confusion with non-similar object category, and BG - confusion with background. The MSS approach is shown to significantly reduce errors due to poor localization.}
 \label{fig:resFPanalysis}
\end{figure*}

\textbf{Relationship between scale-variation, dataset size, and MSS benefit}: Our experiments showed the MSS method to significantly impact performance on some object classes by up to 7 AP points (e.g.  `bottle' and `dining table' classes). Overall, 12 out of the 20 object categories benefit from the MSS approach, specifically on challenging object instances (i.e. small objects) and in terms of localization quality. Furthermore, overall mAP is improved with the MSS approach as shown in Table \ref{tab:restable}. Nonetheless, certain object categories do not benefit from incorporation of the multi-scale reasoning. As the reason for this is not immediately clear, we further study it next. A closer inspection of the scale distribution of the different classes reveals some insight, as shown in Fig. \ref{fig:HOGscaledstbn}. First, a difference between HOG and CNN features is observed. Because CNN features are more scale-sensitive than HOG, this translates into smaller performance gains due to multi-scale context. Employing HOG on the other hand results in large gains consistently and across all object categories. A second observation is that some classes in the PASCAL VOC dataset exhibit smaller variation in scale. This limits the benefits due to incorporation of multi-scale context, and results in smaller AP improvement. If a certain object class exhibits smaller scale variation in the test set, the contextual cues will be less beneficial, which implies the results are influenced by the object statistics in the test set. Finally, we wish to analyze the role of dataset size on the variation in performance. Because the multi-scale templates require scale-specific instances, a small number of instances in the dataset (even with data augmentation) could lead to sub-optimal learning and consequent reduction in performance gains. The importance of sufficient training instances for training each of the scale-specific MSS template is verified in Fig. \ref{fig:datasize_train}. As shown in Fig. \ref{fig:datasize_train}, classes with low detection AP improvement also contain a small number of objects in multiple image scales. In Fig. \ref{fig:datasize_train}, low instance count is defined as a value under the average number of instances per scale bin across all object categories. Together with the observations in Fig. \ref{fig:HOGscaledstbn} regarding limited scale variability and insufficient training data explain why detection of certain classes, such as `bottle', `aeroplane', `dining-table', and `sofa', greatly benefit from the multi-scale context framework, and some classes do not (mainly `boat' and `bird' which contain small scale variability as shown in Fig. \ref{fig:HOGscaledstbn})). As will be shown next, the MSS approach significantly improves localization quality across all object categories.

\textbf{Localization quality}: Fig. \ref{fig:resFPanalysis} demonstrates improved localization due to incorporation of contextual cues across scales. The improvement is consistent over all types of object categories (clustered into three super-classes), including furniture, vehicles, and animals. This type of analysis is encouraging, as CNN-based object detectors are known to suffer from in-accurate localization. Our approach demonstrates the benefit on localization due to explicit incorporation of multi-scale features. This is intuitive, as the existence of certain feature responses at some scales can assist in better localization at another scale.

\textbf{Context statistics}: Training MSS models places discriminative value on each multi-scale cue. Next, we aim to understand how important are such cues in the learning process. For each class, features were divided into two: 1) Features found in the best-fit scale corresponding to the same features that would be employed if a single-scale template (referred to as `in-scale' features), and 2) `out-of-scale' features which a placed outside of the best-fit scale. The learned parameters, $w$, can be decomposed to positive and negative valued entries as $w = w^{+}+w^{-}$. Indices with higher absolute value correspond to locations in the feature space which provide large discriminative value. Single-scale model training involves only `in-scale' features. Furthermore, if `out-of-scale' features provided no benefit, we would expect the majority of the discriminative weight to be placed on the best-fit `in-scale' features only.

\begin{figure*}[!t]
\centering
\begin{tabular}{c}
 \includegraphics[width=6in]{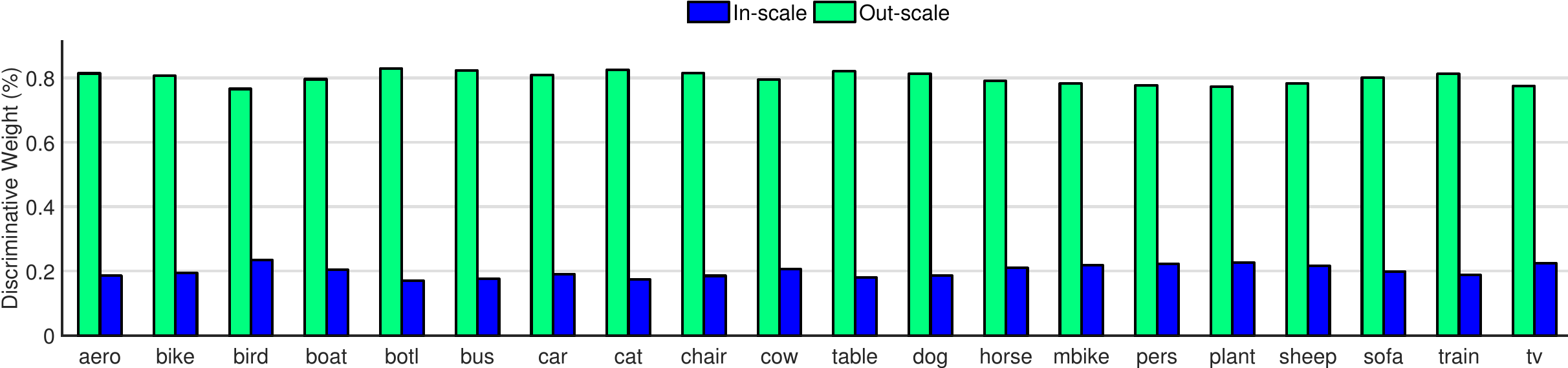}
 \end{tabular}
 \caption{For CNN-based detection at a given scale, how important are out-of-scale context features? See Sec. \ref{sec:virtual} for details.}
 \label{fig:outscaleimportance}
\end{figure*}

\begin{figure*}[!t]
\centering
\begin{tabular}{c}
 \includegraphics[width=6in]{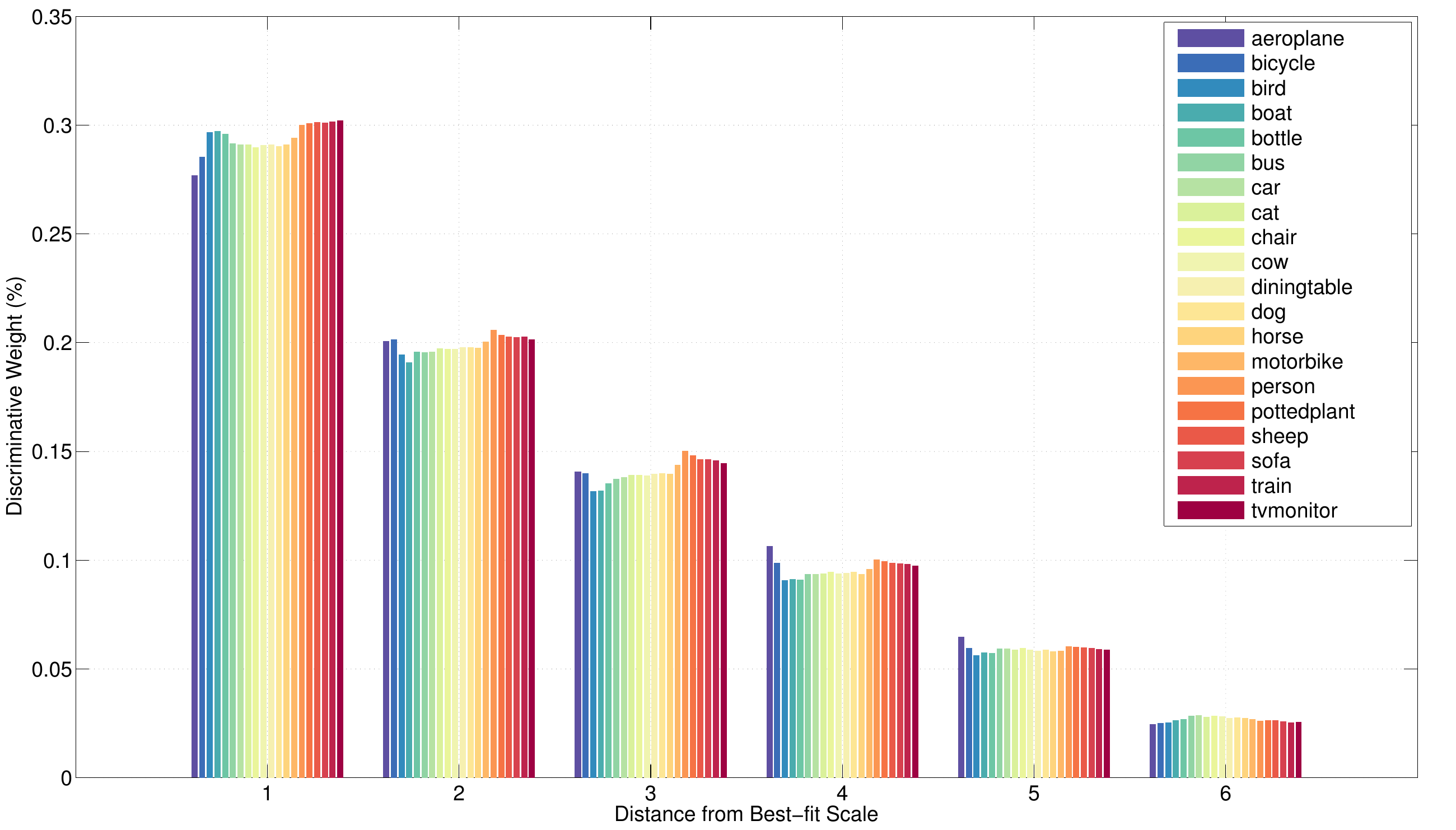}
 \end{tabular}
 \caption{Relative to the best-fit scale, how is discriminative value distributed across pyramid levels? Most of the weight is found within adjacent levels (distance of `1' level away), but the contextual cues are shown to span all levels.}
 \label{fig:scalebestfitdist}
\end{figure*}
  %This allows us to quantify the improvement in detecting and localizing due to incorporation of out-of-scale features. 

By studying the percentage of discriminative weight in $w^{+}$ and its distribution across scales for MSS-CNN, Fig. \ref{fig:outscaleimportance} demonstrates the clear trend of choosing features that are placed outside of the ground truth scale in training. This is a data-driven affirmation of the proposed approach. Although only positive weights shown in Fig. \ref{fig:outscaleimportance}, the trends are similar both over positive weights $w^{+}$ and negative weights $w^{-}$. We can see that context can benefit CNN-detection greatly. %Furthermore, 
 
  %

%\begin{landscape}
\begin{table*}[!t]
  \centering
  \caption{Detection average precision (\%) on VOC 2007 test. Column C shows the number of aspect ratio components. Performance improvement due to incorporation of context and multi-scale reasoning (MSS) with HOG and CNN features are shown. For reference, two other baselines, of a three aspect ratio components single-scale model and region proposal-based approach, are included. Note that the results of \cite{dpmarecnn} for one and three aspect ratio components are using the publicly available code.}
  \resizebox{17.4cm}{!} {

\begin{tabular}{c|c|cccccccccccccccccccc|c}
      & C     & aero  & bike  & bird  & boat  & botl  & bus   & car   & cat   & chair & cow   & table & dog   & horse & mbike & pers  & plant & sheep & sofa  & train & tv    & mAP \\
\hline
HOG   & 1     & 13.05 & 23.54 & 0.80  & 1.70  & 12.85 & 28.91 & 27.38 & 0.68  & 11.31 & 8.89  & 11.04 & 2.68  & 13.52 & 18.49 & 13.05 & 5.60  & 14.58 & 12.19 & 16.28 & 24.48 & 13.05 \\
HOG-MSS & 1     & 21.72 & 33.86 & 10.05 & 1.81  & 12.02 & 22.54 & 40.04 & 24.66 & 13.52 & 10.08 & 20.28 & 13.53 & 32.57 & 23.63 & 23.05 & 7.24  & 18.23 & 22.75 & 24.20 & 33.98 & 20.49 \\
      &       &       &       &       &       &       &       &       &       &       &       &       &       &       &       &       &       &       &       &       &       &  \\
       \hline
            &       &       &       &       &       &       &       &       &       &       &       &       &       &       &       &       &       &       &       &       &       & \textbf{} \\
CNN \cite{dpmarecnn}   & 1     & 33.54 & 55.95 & 24.97 & \textbf{14.24} & \textbf{36.96} & \textbf{44.31} & 52.33 & 40.37 & \textbf{30.07} & 44.56 & 9.09  & 34.47 & 51.26 & 53.39 & 38.66 & \textbf{25.22} & 40.16 & 41.36 & 36.31 & \textbf{57.97} & 38.26 \\
CNN-ours & 1   & 36.68 & \textbf{60.66} & \textbf{33.45} & 13.71 & 17.66 & 44.02 & 58.48 & 49.71 & 25.12 & \textbf{46.32} & 44.08 & 41.47 & \textbf{57.76} & 54.18 & 48.90 & 22.95 & 43.84 & 43.34 & 42.17 & 54.96 & 41.97 \\
CNN-MSS & 1     & \textbf{41.88} & 56.17 & 30.40 & 12.54 & 25.05 & 43.36 & \textbf{60.75} & \textbf{50.27} & 27.68 & 45.41 & \textbf{51.25} & \textbf{41.94} & 55.60 & \textbf{55.71} & \textbf{49.30} & 22.25 & \textbf{43.91} & \textbf{46.22} & \textbf{42.27} & 52.78 & \textbf{42.74} \\
      &       &       &       &       &       &       &       &       &       &       &       &       &       &       &       &       &       &       &       &       &       & \textbf{} \\
      \hline
            &       &       &       &       &       &       &       &       &       &       &       &       &       &       &       &       &       &       &       &       &       & \textbf{} \\
CNN \cite{dpmarecnn}   & 3     & 44.64 & 64.49 & 32.43 & 23.53 & 35.64 & 55.92 & 56.90 & 39.38 & 28.07 & 49.64 & 42.18 & 41.38 & 59.95 & 55.52 & 53.92 & 24.55 & 46.81 & 38.89 & 47.53 & 59.39 & 45.04 \\
R-CNN pool$_5$ \cite{rcnn}  &  -      & 51.8  & 60.2  & 36.4  & 27.8  & 23.2  & 52.8  & 60.6  & 49.2  & 18.3  & 47.8  & 44.3  & 40.8  & 56.6  & 58.7  & 42.4  & 23.4  & 46.1  & 36.7  & 51.3  & 55.7  & 44.2 \\
\hline
    \end{tabular}%
    }
  \label{tab:restable}%
\end{table*}%

\begin{table*}[!t]
  \centering
  \caption{ The table depicts detection average precision (\%) on VOC 2007 test for other methods employing \textbf{part modeling and CNN features}. The results are included for completeness, and meant to be compared with the results in Table \ref{tab:restable}. Our proposed method does not perform any explicit part reasoning. }
  \resizebox{17.4cm}{!} {
\begin{tabular}{c|c|c|cccccccccccccccccccc|c}
      & C  & P  & aero  & bike  & bird  & boat  & botl  & bus   & car   & cat   & chair & cow   & table & dog   & horse & mbike & pers  & plant & sheep & sofa  & train & tv    & mAP \\
\hline
C-DPM \cite{CNNdpm2}& 3 & 8 & 39.7& 59.5& \textbf{35.8}& 24.8 &35.5& 53.7& 48.6& \textbf{46.0} &\textbf{29.2} &\textbf{36.8}& \textbf{45.5}& \textbf{42.0} &57.7 &56.0 &37.4& \textbf{30.1} &31.1& \textbf{50.4} &\textbf{56.1}& 51.6& \textbf{43.4}
\\
Conv-DPM \cite{parts1}& 3 & 9 & \textbf{48.9} &\textbf{67.3}& 25.3 &\textbf{25.1}&\textbf{ 35.7} &\textbf{58.3}& \textbf{60.1} &35.3 &22.7& 36.4& 37.1& 26.9& \textbf{64.9} &\textbf{62.0}& \textbf{47.0} &24.1& \textbf{37.5}& 40.2& 54.1 &\textbf{57.0}& 43.3
\\
\hline
    \end{tabular}%
    }
  \label{tab:restablewithparts}%
\end{table*}%

\begin{table*}[!t]
  \centering
  \caption{Results with fine-tuned features on VOC 2007 test. Our approach uses no region proposals (unlike RCNN), a single aspect ratio model, and only conv$_5$ feature maps. }
  \resizebox{17.4cm}{!} {
    \begin{tabular}{c|c|cccccccccccccccccccc|c}
          & C     & aero  & bike  & bird  & boat  & botl  & bus   & car   & cat   & chair & cow   & table & dog   & horse & mbike & pers  & plant & sheep & sofa  & train & tv    & mAP \\
    \hline
    CNN   & 1     & 41.48 & 62.58 & 36.88 & 16.65 & 22.23 & 48.07 & 61.31 & 50.78 & 29.41 & 49.10 & 47.54 & 45.64 & 62.45 & 58.13 & 50.61 & 25.57 & 48.58 & 48.01 & 44.81 & \textbf{59.53} & 45.47 \\
    CNN-MSS & 1     & 46.68 & 58.09 & 33.83 & 15.48 & \textbf{29.62} & 47.41 & 63.58 & 51.34 & \textbf{31.97} & 48.19 & \textbf{54.71} & \textbf{46.11} & \textbf{60.29} & \textbf{59.66} & \textbf{51.01} & 24.87 & 48.65 & \textbf{50.89} & 44.91 & 57.35 & 46.23 \\
    RCNN pool$_5$ \cite{rcnn} &   -    & \textbf{58.2} & \textbf{63.3} & \textbf{37.9} & \textbf{27.6} & 26.1  & \textbf{54.1} & \textbf{66.9} & \textbf{51.4} & 26.7  & \textbf{55.5} & 43.4  & 43.1  & 57.7  & 59.0    & 45.8  & \textbf{28.1} & \textbf{50.8} & 40.6  & \textbf{53.1} & 56.4  & \textbf{47.3} \\
          &       &       &       &       &       &       &       &       &       &       &       &       &       &       &       &       &       &       &       &       &       &  \\
          \hline
                    &       &       &       &       &       &       &       &       &       &       &       &       &       &       &       &       &       &       &       &       &       &  \\
    RCNN fc$_7$ \cite{rcnn} &    -  & 64.2  & 69.7  & 50.0   & 41.9  & 32.0    & 62.6  & 71.0    & 60.7  & 32.7  & 58.5  & 46.5  & 56.1  & 60.6  & 66.8  & 54.2  & 31.5  & 52.8  & 48.9  & 57.9  & 64.7  & 54.2 \\
    RCNN fc$_7$ BB \cite{rcnn} &    -  & 68.1 & 72.8& 56.8& 43.0& 36.8& 66.3& 74.2& 67.6 &34.4 &63.5 &54.5 &61.2& 69.1& 68.6& 58.7& 33.4& 62.9& 51.1& 62.5& 64.8& 58.5 \\
 \hline
    \end{tabular}%
    }
  \label{tab:withft}%
\end{table*}%
%\end{landscape}

\begin{figure}[!t]
\centering
\begin{tabular}{cc}
\includegraphics[width=2in]{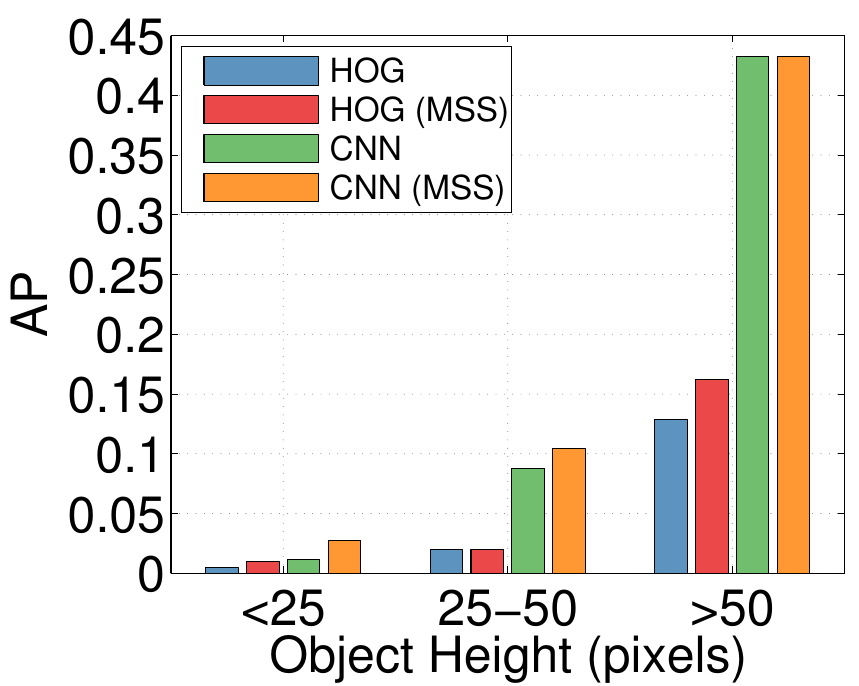}

 \end{tabular}
 \caption{Improvement in performance for different object sizes. The largest gains due to incorporating the MSS approach are seen on smaller objects, which include more relevant contextual information throughout the multi-scale features.}
 \label{fig:resHiLowres}
\end{figure}

A further breakdown of this information is visualized in Fig. \ref{fig:scalebestfitdist}. Here, it is shown that most of the features selected outside of the best-fit scale are located in the adjacent scale (a distance of `1' pyramid level away), which is to be expected. Nonetheless, the MSS models consistently select features at more remote pyramid levels, even up to more than an octave away. This analysis suggests that CNN-based approaches can greatly benefit from careful multi-scale and contextual reasoning, which is not done in most existing approaches for object detection. Simple pooling over both adjacent and remote scales is shown to greatly assist in detection, as shown in Fig. \ref{fig:scalebestfitdist}. Interestingly, a spike at certain remote scales is clearly seen with some categories, such as `aeroplane', `bicycle', and `person'. This observation can be better understood by inspecting the template visualization in Fig. \ref{fig:vismodelspascal}. For `aeroplane', many of the scales contain informative contextual information as shown in Fig. \ref{fig:vismodelspascal}, from wings to other aeroplanes. For `bicycle', a rider may be found at a further scale. It can also be clearly observed how classes which MSS benefits least (`bird' and `boat') have the smallest discriminative value placed in other scales out of all object categories. In these classes, contextual information is not selected as much.

\textbf{Performance breakdown by scale}: As shown in Fig. \ref{fig:resHiLowres}, most gains in detection performance with CNN-features come from detection of smaller objects (50 pixels and less in height). This is intuitive, as such objects can benefit from incorporation of contextual cues at other scales.

\textbf{Comparison with state-of-the-art}: The main emphasis in this work is in analysis on modeling multi-scale context and its applications to efficient object detection and localization with deep features. The analysis framework was used to study scale importance, impact of dataset properties, and performance under varying object class and size settings. On PASCAL VOC, certain object classes greatly benefited from the proposed approach in detection, all of the 20 classes benefited in localization quality, and insights were made regarding challenging cases for the MSS approach. By employing only conv$_5$ feature maps, the method is efficient (requiring a single forward pass for each image scale) and have a low memory impact (no fully connected layers which contain most of the network parameters). As a reference, we provide absolute performance to other related research studies in Tables \ref{tab:restable}, \ref{tab:restablewithparts}, and \ref{tab:withft} with different experimental settings. 

For a fair comparison with a baseline, we closely followed Girshick \etal \cite{dpmarecnn} in the deep feature pyramid extraction throughout the experiments. Overall, with a single aspect ratio model, our analysis results in a significant improvement of $4.48$ mAP over the results of \cite{dpmarecnn}, from $38.26$ mAP (obtained by the available implementation of \cite{dpmarecnn}) to $42.74$. We observed model size to be a crucial parameter, and increasing it results in improvement of the baseline to $41.97$ mAP. Large gains in detection performance are shown for HOG, with an mAP increase of over 7 points. As discussed previously, the MSS approach has less impact on objects with little scale variation. Furthermore, as multi-scale templates require scale-specific instances, a small number of instances in the dataset (even with data augmentation), leads to sub-optimal learning and reduced performance gains. On the other hand, certain classes (e.g. `aeroplane', `car', `table', and `sofa') show large gains in performance. As the method in \cite{dpmarecnn} employs no contextual reasoning, a further gain is obtained by the multi-scale reasoning in overall mAP.

As a reference, although not the main focus of this study, the results of \cite{dpmarecnn} with three aspect ratios are shown, which has an overall $6.78$ points improvement up to $45.04$ AP, improving over R-CNN in performance with the same convolutional feature maps. The improvement due to multiple aspect ratio components is an orthogonal improvement to MSS as context cues can be incorporated into each of the components. Furthermore, note that unlike R-CNN, \cite{dpmarecnn} and our study does not involve a region proposal mechanism and per-region forward pass through the network (either through the whole network or just through the fully connected layers), which is computationally costly. The CNN-MSS approach ($42.74$ mAP) performs similarly to other recently proposed approaches of Wan \etal \cite{parts1} and Savalle \etal \cite{CNNdpm2} employing multiple aspect ratio components, CNN feature pyramids, and explicit part reasoning. The best relevant results is achieved with R-CNN, fine-tuning, multiple fully connected layers (fc$_7$), and bounding-box (BB) regression at $58.5$ mAP. Compared to R-CNN, the proposed approach is significantly more efficient in memory and computational cost. Furthermore, MSS learns scale-specific appearance and localization models while R-CNN does not. Results are shown both for no fine-tuning and with fine-tuning. R-CNN with the same convolutional features is outperformed on some classes where region proposals are weak. The results post fine-tuning shown in Table \ref{tab:withft} demonstrate a consistent improvement. This is expected, as fine-tuning is mostly focused on improving local region representation.

\textbf{Run-time speed}: The computational speed is bound by two main factors, the feature pyramid extraction time and the model evaluation (either single-scale or MSS). The feature computation step (a 7 scale deep feature pyramid) is identical for the baseline and the MSS approach, running at $\sim0.4$ seconds per image on PASCAL with a Titan X GPU. For the baseline, scoring a window $p_s$ using the features $\phi(p_s) \in \mathbb{R}^d$ involves $d$ operations, which is repeated over $S$ scales ($S \times d$). For a given image location, evaluation with the MSS detector involves $S$ models and an increase of the computational cost by a factor of $S$, to ($S \times S \times d$). In the current CPU implementation, the run-time of the MSS evaluation takes $\sim0.7$ seconds per image. In the future, feature selection could potentially reduce the computational complexity of the detector evaluation for further speed gains.

\begin{figure}[!t]
\centering
\begin{tabular}{c}
\includegraphics[width=2.5in]{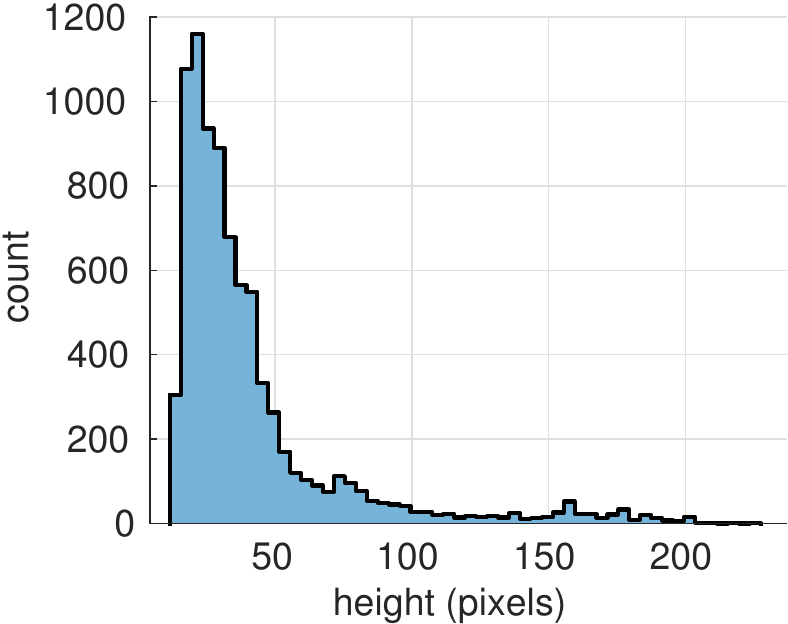} \end{tabular}
 \caption{Height distribution in the highway vehicles dataset.}
 \label{fig:hwystats}
\end{figure}

\subsection{Results on Highway Vehicles}
The PASCAL VOC 2007 dataset was used for developing the MSS approach and providing analysis in terms of impact of dataset properties, error types and localization quality, generalization to different object types, and sensitivity to object scale. In order to further test the performance of the proposed approach and understand its benefits, we employ a multi-view highway dataset captured using front and rear mounted cameras on a moving vehicle platform \cite{jacmikcvpr}. The highway settings are relevant as objects undergo large variation in scale as they enter and leave the scene. Furthermore, because the PASCAL VOC 2007 dataset targets generic object detection, it only contains a handful of images in settings similar to highway settings. The highway vehicles dataset is composed of a total of 1550 images containing 8295 objects. All truncated vehicles are also included in the evaluation. Object occlusion state have also been annotated in order to study performance under occlusion. The object height distribution is depicted in Fig. \ref{fig:hwystats}, showing large variation. 

\begin{figure}[!t]
\centering
\begin{tabular}{cc}
\includegraphics[width=1.5in]{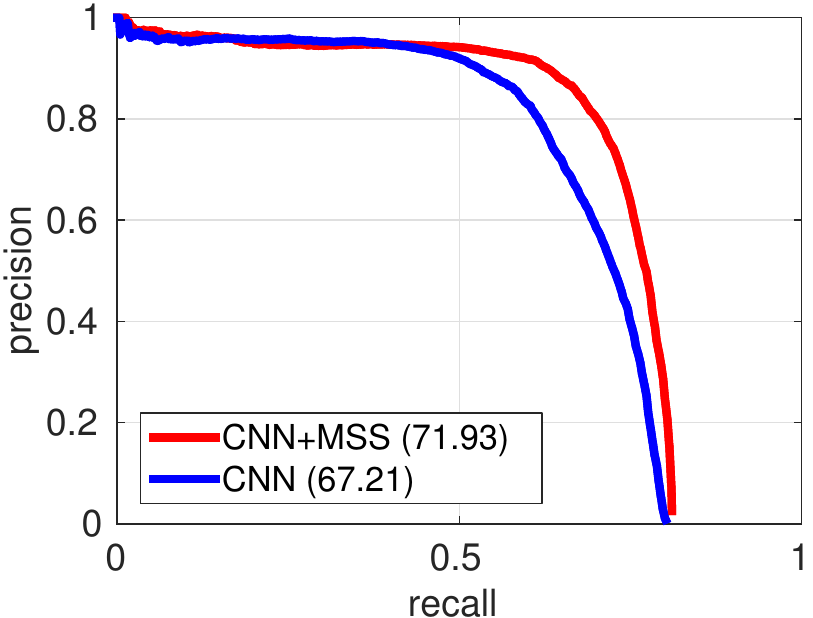} & \includegraphics[width=1.5in]{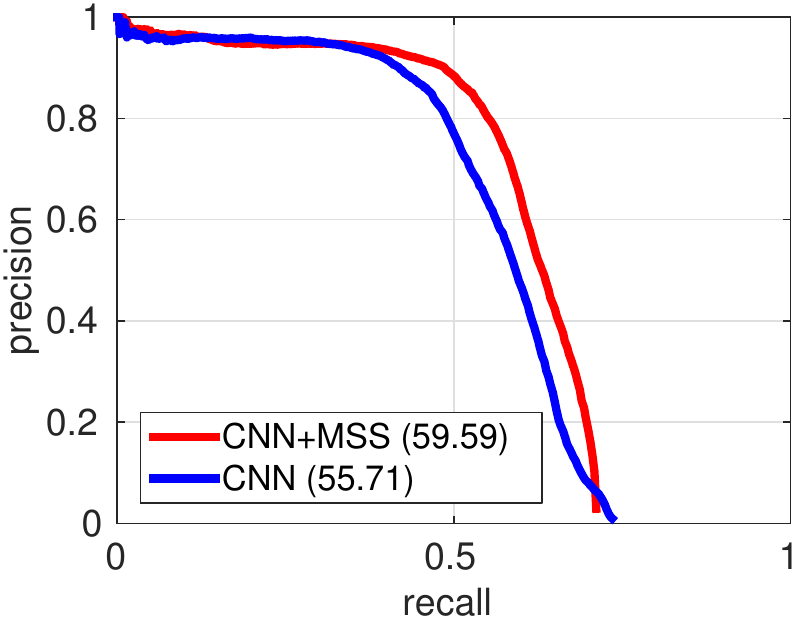} 
\\
(a) Easy - No Occlusion & (b) Hard - With Occlusion 
\\
\includegraphics[width=1.5in]{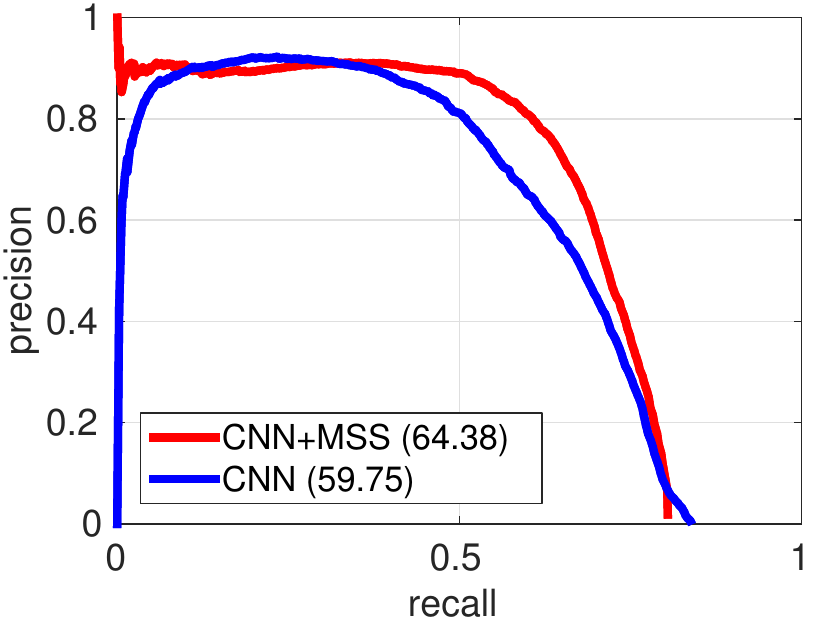} & \includegraphics[width=1.5in]{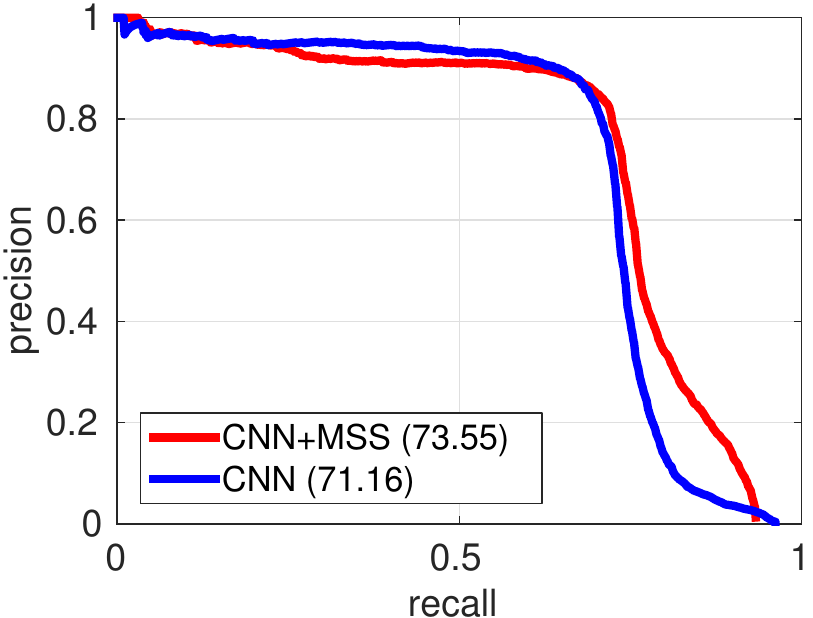} 
\\
(c) Height - 20-40 Pixels & (d) Height - Above 40 Pixels
 \end{tabular}
 \caption{Results for vehicle detection on highway settings with different evaluation procedures.}
 \label{fig:hwy}
\end{figure}

\begin{figure}[!t]
\centering
\begin{tabular}{cc}
\includegraphics[width=1.31in]{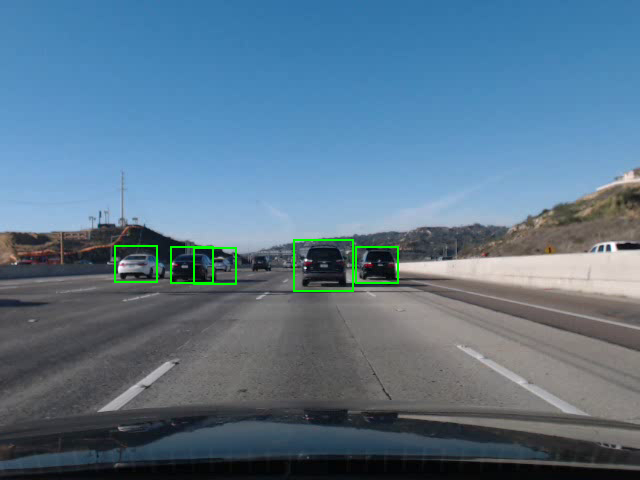} & \includegraphics[width=1.31in]{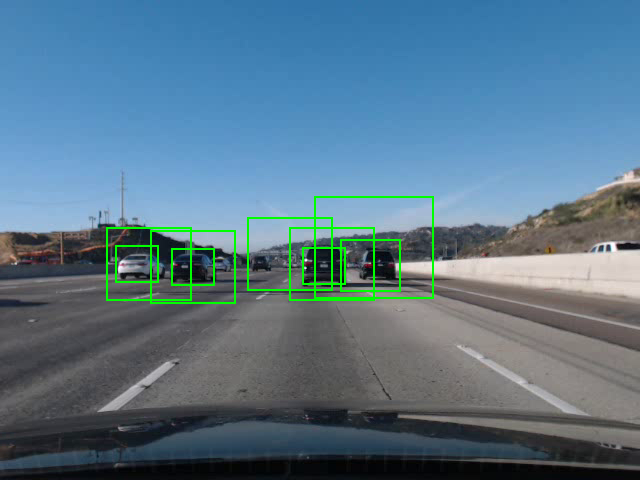} 
\\
\includegraphics[width=1.31in]{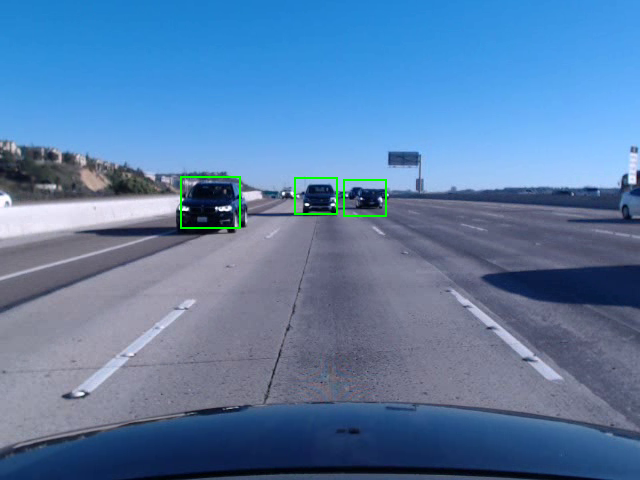} & \includegraphics[width=1.31in]{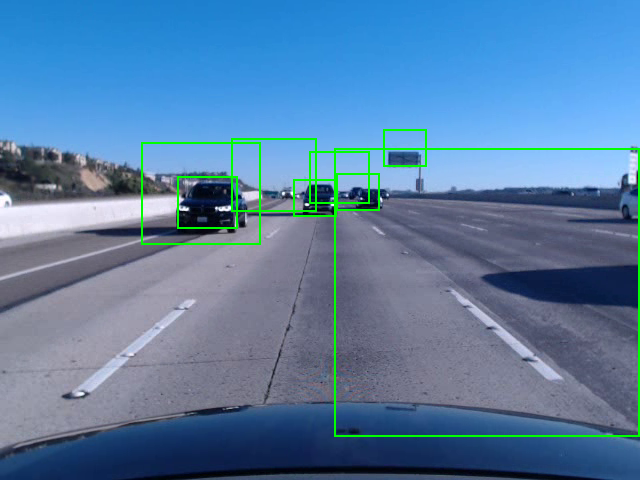} 
\\
\includegraphics[width=1.31in]{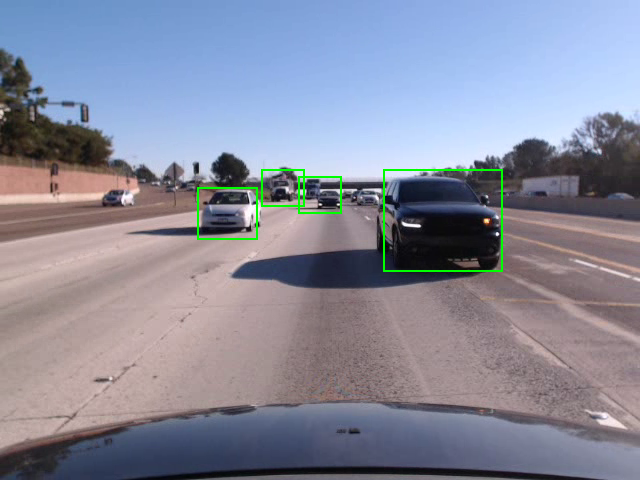} & \includegraphics[width=1.31in]{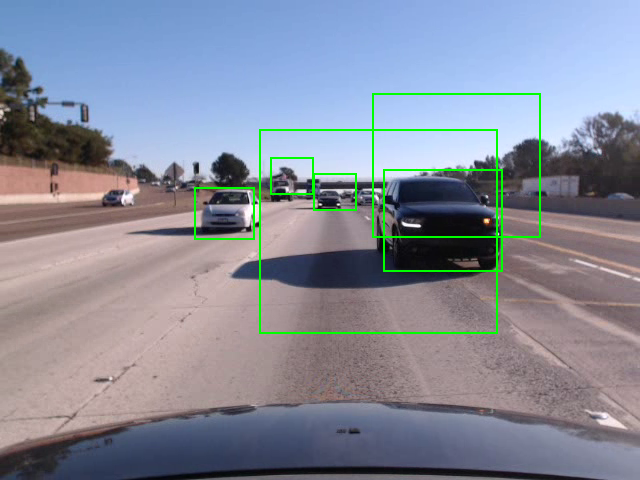} 
\\
\includegraphics[width=1.31in]{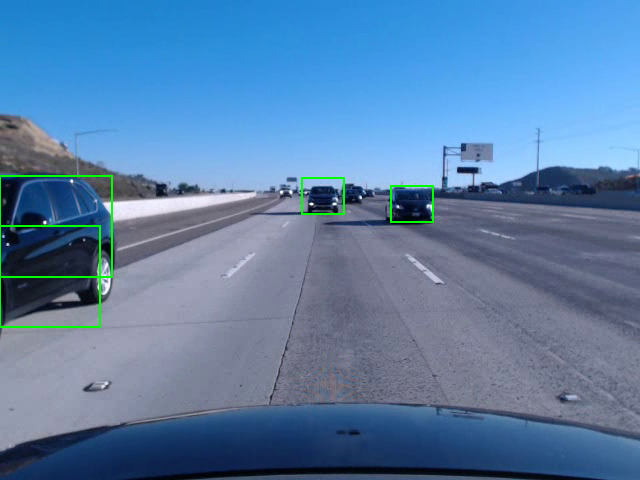} & \includegraphics[width=1.31in]{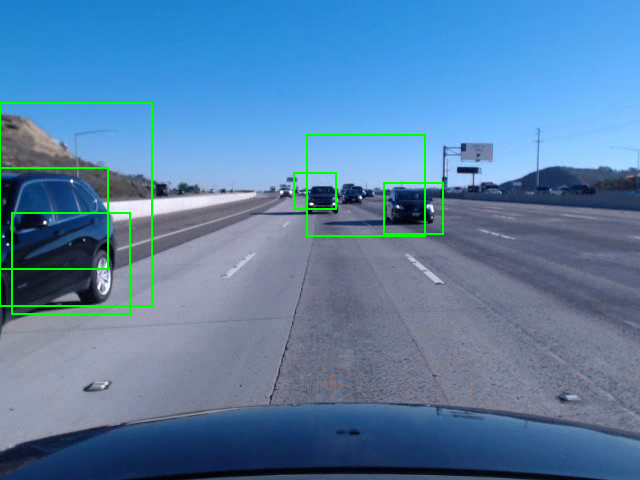} 
\\
\includegraphics[width=1.31in]{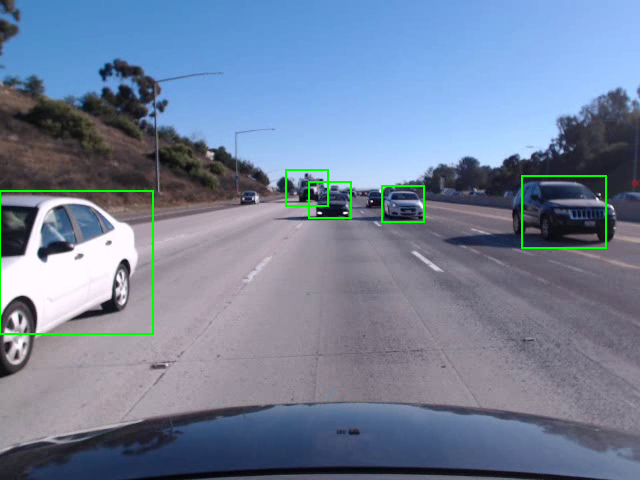} & \includegraphics[width=1.31in]{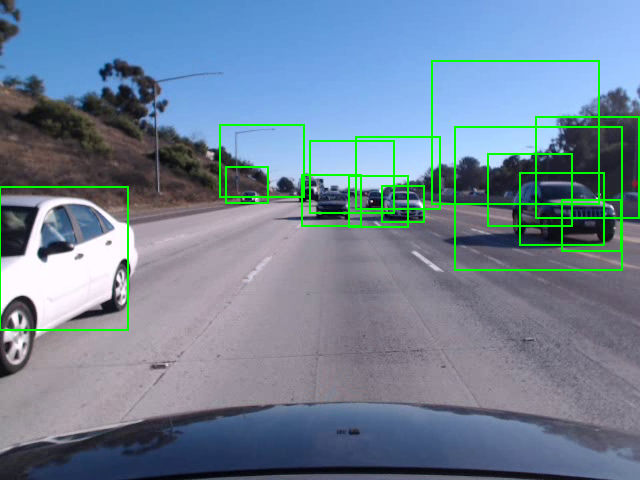} 
\\
\includegraphics[width=1.31in]{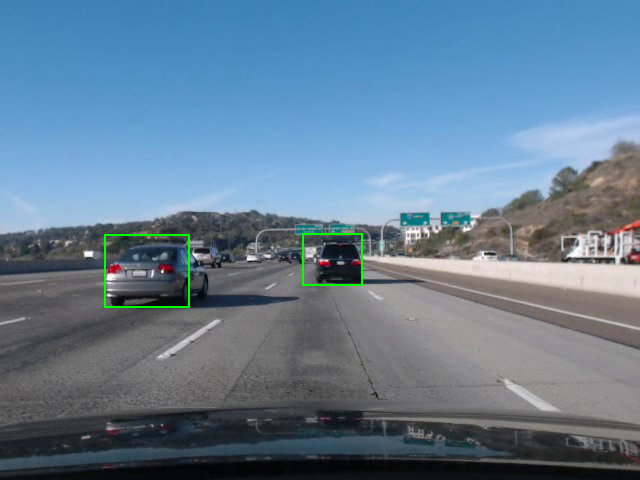} & \includegraphics[width=1.31in]{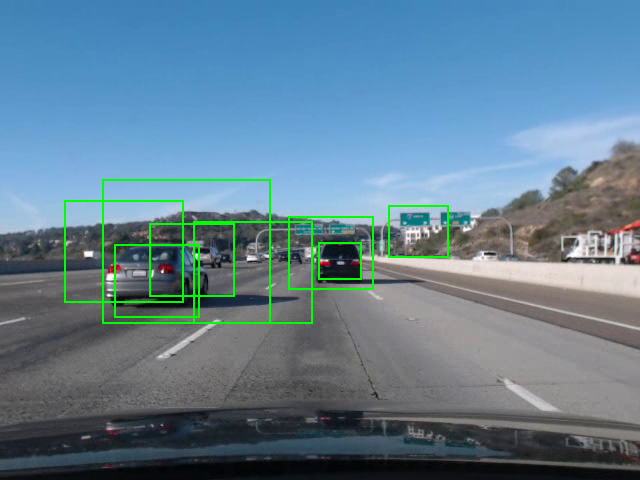} 
\\
%\includegraphics[width=1.5in]{revise3/lisaaims0.73/1266_MSS_.png} & \includegraphics[width=1.5in]{revise3/lisaaims0.73/1266_TP_.png} 
%\\
(a) CNN-MSS (proposed) & (b) CNN (baseline)
 \end{tabular}
 \caption{Results for vehicle detection on highway settings at a fixed recall rate. Observe how the MSS approach better reasons over multi-scale responses, allowing for higher precision at the same recall rate and better localization compared to the single-scale CNN, which employs independent scoring at each scale and relies on NMS alone for resolving multi-scale responses.}
 \label{fig:hwyim}
\end{figure}

The results for vehicle detection are shown in Fig. \ref{fig:hwy}. When occluded objects are excluded, the MSS approach results in a significant improvement of $4.72$ AP points over the baseline. With the inclusion of occluded objects, the improvement is consistent at $3.88$ AP points. On this dataset, a main improvement is in detecting smaller objects and better resolving multiple detection boxes, as shown in Fig. \ref{fig:hwy}(c). By observing the curves in Fig. \ref{fig:hwy}, we can see how the MSS approach maintains precision at a higher recall over the baseline. This is due to the improved multi-scale reasoning. While the baseline scores objects based on local information and therefore relies on the heuristic NMS alone to resolve responses at nearby locations and multiple scales, the MSS approach can better reason over responses in different scales. This can be clearly seen in the example images in Fig. \ref{fig:hwyim}, where detection results are shown for both the MSS and the single scale baseline at a fixed recall rate. Fig. \ref{fig:hwyim} also shows cases where false positives are reduced due to contextual information available at multiple scales.

\section{Concluding Remarks}

In this paper, the role of multi-scale context in object detection with deep features was studied. An efficient framework for analysis of multi-scale contextual reasoning was proposed and studied on the PASCAL object detection benchmark and a highway vehicles dataset. Because the proposed approach operates on scale volumes, learns scale-specific models, and infers a localization label, it was shown to result in more robust detection and localization of objects. Visualization and feature selection analysis demonstrated how discriminative learning strongly favor multi-scale cues when these are present in training, both in adjacent and remote image scales. Comparative analysis evaluated generalization of the proposed approach for different feature types and dataset settings. As current state-of-the-art object detectors emphasize local region feature pooling in detection, the insights in this study can be used to train better CNN-based object detectors. In the future, the insights from this study will be used in order to design better end-to-end contextual, multi-scale detection frameworks.  

\section{Acknowledgments}

We thank the reviewers and editors for their comments while preparing the manuscript, and thank our colleagues at the Computer Vision and Robotics Research Laboratory for their assistance. We also thank Zhuowen Tu for helpful discussions. 

{\small
\bibliographystyle{ieee}
\bibliography{IEEEexample}

\begin{thebibliography}{10}\itemsep=-1pt

\bibitem{peds100rodrigo}
R.~Benenson, M.~Mathias, R.~Timofte, and L.~V. Gool.
\newblock Pedestrian detection at 100 frames per second.
\newblock In {\em IEEE Conf. Computer Vision and Pattern Recognition}, 2012.

\bibitem{ssvmlocalization}
M.~Blaschko and C.~Lampert.
\newblock Learning to localize objects with structured output regression.
\newblock In {\em European Conf. Computer Vision}, 2008.

\bibitem{branson}
S.~Branson, O.~Beijbom, and S.~Belongie.
\newblock Efficient large-scale structured learning.
\newblock In {\em IEEE Conf. Computer Vision and Pattern Recognition}, 2013.

\bibitem{vggcnn}
K.~Chatfield, K.~Simonyan, A.~Vedaldi, and A.~Zisserman.
\newblock Return of the devil in the details: Delving deep into convolutional
  networks.
\newblock In {\em British Machine Vision Conf.}, 2014.

\bibitem{moco}
G.~Chen, Y.~Ding, J.~Xiao, and T.~X. Han.
\newblock Detection evolution with multi-order contextual co-occurrence.
\newblock In {\em IEEE Conf. Computer Vision and Pattern Recognition}, 2013.

\bibitem{nmslayout}
C.~Desai, D.~Ramanan, and C.~C. Fowlkes.
\newblock Discriminative models for multi-class object layout.
\newblock {\em Intl. Journal Computer Vision}, 95(1):1--12, 2011.

\bibitem{contextboost}
Y.~Ding and J.~Xiao.
\newblock Contextual boost for pedestrian detection.
\newblock In {\em IEEE Conf. Computer Vision and Pattern Recognition}, 2012.

\bibitem{DollarPAMI14pyramids}
P.~Doll{\'a}r, R.~Appel, S.~Belongie, and P.~Perona.
\newblock Fast feature pyramids for object detection.
\newblock {\em IEEE Trans. Pattern Analysis and Machine Intelligence},
  36(8):1532--1545, 2014.

\bibitem{mscnn}
D.~Eigen, C.~Puhrsch, and R.~Fergus.
\newblock Depth map prediction from a single image using a multi-scale deep
  network.
\newblock In {\em Neural Information Processing Systems}, 2014.

\bibitem{pascalvoc}
M.~Everingham, L.~V. Gool, C.~K.~I. Williams, J.~Winn, and A.~Zisserman.
\newblock The {PASCAL} visual object classes {(VOC)} challenge.
\newblock {\em Intl. Journal Computer Vision}, 88(2):303--338, 2010.

\bibitem{yannmssseg}
C.~Farabet, C.~Couprie, L.~Najman, and Y.~LeCun.
\newblock Learning hierarchical features for scene labeling.
\newblock {\em IEEE Trans. Pattern Analysis and Machine Intelligence},
  35(8):1915--1929, 2013.

\bibitem{dpmPAMI}
P.~Felzenszwalb, R.~Girshick, D.~McAllester, and D.~Ramanan.
\newblock Object detection with discriminatively trained part-based models.
\newblock {\em IEEE Trans. Pattern Analysis and Machine Intelligence},
  32(9):1627--1645, 2010.

\bibitem{fukushima1980neocognitron}
K.~Fukushima.
\newblock Neocognitron: A self-organizing neural network model for a mechanism
  of pattern recognition unaffected by shift in position.
\newblock {\em Biological cybernetics}, 36(4):193--202, 1980.

\bibitem{Geiger2013IJRR}
A.~Geiger, P.~Lenz, C.~Stiller, and R.~Urtasun.
\newblock Vision meets robotics: The {KITTI} dataset.
\newblock {\em International Journal of Robotics Research}, 32(11):1231--1237,
  2013.

\bibitem{girshick15fastrcnn}
R.~Girshick.
\newblock Fast {R-CNN}.
\newblock In {\em IEEE Intl. Conf. on Computer Vision}, 2015.

\bibitem{rcnn}
R.~Girshick, J.~Donahue, T.~Darrell, and J.~Malik.
\newblock Rich feature hierarchies for accurate object detection and semantic
  segmentation.
\newblock In {\em IEEE Conf. Computer Vision and Pattern Recognition}, 2014.

\bibitem{dpmarecnn}
R.~Girshick, F.~Iandola, T.~Darrell, and J.~Malik.
\newblock Deformable part models are convolutional neural networks.
\newblock In {\em IEEE Conf. Computer Vision and Pattern Recognition}, 2015.

\bibitem{sppnet}
K.~He, X.~Zhang, S.~Ren, and J.~Sun.
\newblock Spatial pyramid pooling in deep convolutional networks for visual
  recognition.
\newblock In {\em European Conf. Computer Vision}, 2014.

\bibitem{Hoai20141523}
M.~Hoai, L.~Torresani, F.~D. la~Torre, and C.~Rother.
\newblock Learning discriminative localization from weakly labeled data.
\newblock {\em Pattern Recognition}, 47(3):1523--1534, 2014.

\bibitem{hoiem}
D.~Hoiem, Y.~Chodpathumwan, and Q.~Dai.
\newblock Diagnosing error in object detectors.
\newblock In {\em European Conf. Computer Vision}, 2012.

\bibitem{context1}
D.~Hoiem, A.~Efros, and M.~Hebert.
\newblock Putting objects in perspective.
\newblock In {\em IEEE Conf. Computer Vision and Pattern Recognition}. 2006.

\bibitem{joachims}
T.~Joachims, T.~Finley, and C.-N. Yu.
\newblock Cutting-plane training of structural {SVM}s.
\newblock {\em Machine Learning}, 77(1):27--59, 2009.

\bibitem{jacmikcvpr}
M.~S. Kristoffersen, J.~V. Dueholm, R.~K. Satzoda, M.~M. Trivedi, A.~Mogelmose,
  and T.~B. Moeslund.
\newblock Towards semantic understanding of surrounding vehicular maneuvers: A
  panoramic vision-based framework for real-world highway studies.
\newblock In {\em IEEE Conf. Computer Vision and Pattern Recognition
  Workshops}, 2016.

\bibitem{Krizhevsky}
A.~Krizhevsky, I.~Sutskever, and G.~Hinton.
\newblock {ImageNet} classification with deep convolutional neural networks.
\newblock In {\em Neural Information Processing Systems}, 2012.

\bibitem{LacosteJulien2013}
S.~Lacoste-Julien, M.~Jaggi, M.~Schmidt, and P.~Pletscher.
\newblock Block-coordinate {F}rank-{W}olfe optimization for structural {SVMs}.
\newblock In {\em Intl. Conf. Machine Learning}, 2013.

\bibitem{lecun1989backpropagation}
Y.~LeCun, B.~Boser, J.~S. Denker, D.~Henderson, R.~E. Howard, W.~Hubbard, and
  L.~D. Jackel.
\newblock Backpropagation applied to handwritten zip code recognition.
\newblock {\em Neural computation}, 1(4):541--551, 1989.

\bibitem{li2014integrating}
B.~Li, T.~Wu, and S.-C. Zhu.
\newblock Integrating context and occlusion for car detection by hierarchical
  {And-Or} model.
\newblock In {\em European Conf. Computer Vision}. 2014.

\bibitem{regionletssgm}
C.~Long, X.~Wang, G.~Hua, M.~Yang, and Y.~Lin.
\newblock Accurate object detection with location relaxation and regionlets
  relocalization.
\newblock In {\em Asian Conf. Computer Vision}, 2014.

\bibitem{subcat}
E.~Ohn-Bar and M.~M. Trivedi.
\newblock Fast and robust object detection using visual subcategories.
\newblock In {\em CVPRW}, 2014.

\bibitem{ohnbarTITS15}
E.~Ohn-Bar and M.~M. Trivedi.
\newblock Learning to detect vehicles by clustering appearance patterns.
\newblock {\em IEEE Trans. Intelligent Transportation Systems},
  16(5):2511--2521, 2015.

\bibitem{Osaku201560}
D.~Osaku, R.~Nakamura, L.~Pereira, R.~Pisani, A.~Levada, F.~Cappabianco,
  A.~Falcão, and J.~P. Papa.
\newblock Improving land cover classification through contextual-based
  optimum-path forest.
\newblock {\em Information Sciences}, 324:60--87, 2015.

\bibitem{parkmultires}
D.~Park, D.~Ramanan, and C.~Fowlkes.
\newblock Multiresolution models for object detection.
\newblock In {\em European Conf. Computer Vision}, 2010.

\bibitem{holdingback}
B.~Pepik, R.~Benenson, T.~Ritschel, and B.~Schiele.
\newblock What is holding back convnets for detection?
\newblock In {\em German Conf. Pattern Recognition}, 2015.

\bibitem{fixedpt}
L.~Quannan, J.~Wang, Z.~Tu, and D.~P. Wipf.
\newblock Fixed-point model for structured labeling.
\newblock In {\em Intl. Conf. Machine Learning}, 2013.

\bibitem{rakesh}
R.~N. Rajaram, E.~Ohn-Bar, and M.~M. Trivedi.
\newblock Looking at pedestrians at different scales: A multi-resolution
  approach and evaluations.
\newblock In {\em IEEE Trans. Intelligent Transportation Systems}, 2016.

\bibitem{renNIPS15fasterrcnn}
S.~Ren, K.~He, R.~Girshick, and J.~Sun.
\newblock Faster {R-CNN}: Towards real-time object detection with region
  proposal networks.
\newblock In {\em Neural Information Processing Systems}, 2015.

\bibitem{rumelhart1988learning}
D.~E. Rumelhart, G.~E. Hinton, and R.~J. Williams.
\newblock Learning representations by back-propagating errors.
\newblock {\em Cognitive modeling}, 5:3.

\bibitem{rumelhart1985learning}
D.~E. Rumelhart, G.~E. Hinton, and R.~J. Williams.
\newblock Learning internal representations by error propagation.
\newblock Technical report, DTIC Document, 1985.

\bibitem{visualphrase}
M.~A. Sadeghi and A.~Farhadi.
\newblock Recognition using visual phrases.
\newblock In {\em IEEE Conf. Computer Vision and Pattern Recognition}, 2011.

\bibitem{30hzdpm}
M.~A. Sadeghi and D.~Forsyth.
\newblock 30{Hz} object detection with {DPM} {V5}.
\newblock In {\em ECCV}, 2014.

\bibitem{CNNdpm2}
P.-A. Savalle, S.~Tsogkas, G.~Papandreou, and I.~Kokkinos.
\newblock Deformable part models with cnn features.
\newblock In {\em ECCVW}, 2014.

\bibitem{overfeat}
P.~Sermanet, D.~Eigen, X.~Zhang, M.~Mathieu, R.~Fergus, and Y.~LeCun.
\newblock Overfeat: Integrated recognition, localization and detection using
  convolutional networks.
\newblock In {\em Intl. Conf. Learning Representations}, 2014.

\bibitem{cvpr13peds}
P.~Sermanet, K.~Kavukcuoglu, S.~Chintala, and Y.~LeCun.
\newblock Pedestrian detection with unsupervised multi-stage feature learning.
\newblock In {\em IEEE Conf. Computer Vision and Pattern Recognition}, 2013.

\bibitem{sermanet2011traffic}
P.~Sermanet and Y.~LeCun.
\newblock Traffic sign recognition with multi-scale convolutional networks.
\newblock In {\em Intl. Joint Conf. Neural Networks}, 2011.

\bibitem{Shi2016448}
B.~Shi, X.~Bai, and C.~Yao.
\newblock Script identification in the wild via discriminative convolutional
  neural network.
\newblock {\em Pattern Recognition}, 52:448--458, 2016.

\bibitem{vgg}
K.~Simonyan and A.~Zisserman.
\newblock Very deep convolutional networks for large-scale image recognition.
\newblock In {\em Intl. Conf. Learning Representations}, 2015.

\bibitem{slidingCNN}
C.~Szegedy, A.~Toshev, and D.~Erhan.
\newblock Deep neural networks for object detection.
\newblock In {\em Neural Information Processing Systems}, 2013.

\bibitem{tuAuto}
Z.~Tu and X.~Bai.
\newblock Auto-context and its application to high-level vision tasks and {3D}
  brain image segmentation.
\newblock {\em IEEE Trans. Pattern Analysis and Machine Intelligence}, 32(10),
  2010.

\bibitem{selsearch}
J.~R.~R. Uijlings, K.~E.~A. van~de Sande, T.~Gevers, and A.~W.~M. Smeulders.
\newblock Selective search for object recognition.
\newblock {\em Intl. Journal Computer Vision}, 104(2):154--171, 2013.

\bibitem{vedaldi08vlfeat}
A.~Vedaldi and B.~Fulkerson.
\newblock {VLFeat}: An open and portable library of computer vision algorithms.
\newblock \url{http://www.vlfeat.org/}, 2008.

\bibitem{parts1}
L.~Wan, D.~Eigen, and R.~Fergus.
\newblock End-to-end integration of a convolutional network, deformable parts
  model and non-maximum suppression.
\newblock In {\em IEEE Conf. Computer Vision and Pattern Recognition}, 2015.

\bibitem{hed}
S.~Xie and Z.~Tu.
\newblock Holistically-nested edge detection.
\newblock In {\em IEEE Intl. Conf. on Computer Vision}, 2015.

\bibitem{ccf}
B.~Yang, J.~Yan, Z.~Lei, and S.~Z. Li.
\newblock Convolutional channel features.
\newblock In {\em IEEE Intl. Conf. on Computer Vision}, 2015.

\bibitem{netvis}
M.~D. Zeiler and R.~Fergus.
\newblock Visualizing and understanding convolutional networks.
\newblock In {\em European Conf. Computer Vision}, 2014.

\bibitem{dpmbird}
N.~Zhang, R.~Farrell, F.~Iandola, and T.~Darrell.
\newblock Deformable part descriptors for fine-grained recognition and
  attribute prediction.
\newblock In {\em IEEE Intl. Conf. on Computer Vision}, 2013.

\bibitem{resolutionSpeed}
W.~Zhang, G.~Zelinsky, and D.~Samaras.
\newblock Real-time accurate object detection using multiple resolutions.
\newblock In {\em IEEE Intl. Conf. on Computer Vision}, 2007.

\bibitem{Zuo:2015:EBD:2796563.2796624}
Z.~Zuo, G.~Wang, B.~Shuai, L.~Zhao, and Q.~Yang.
\newblock Exemplar based deep discriminative and shareable feature learning for
  scene image classification.
\newblock {\em Pattern Recognition}, 48(10):3004--3015, 2015.

\end{thebibliography}
}

\end{document}